%% file: paper.tex
\documentclass[10pt,twocolumn,letterpaper]{article}

\usepackage{wacv}
\makeatletter
\@namedef{ver@everyshi.sty}{}
\makeatother
\usepackage{tikz}
\usepackage{times}
\usepackage{epsfig}
\usepackage{graphicx}
\usepackage{amsmath}
\usepackage{amssymb}
\usepackage{booktabs}
\usepackage{multirow}
\usepackage{makecell}
\usepackage{float}
\usepackage[accsupp]{axessibility}

\usepackage{graphicx} 
\usepackage{caption,subcaption}

\usepackage[utf8x]{inputenc}
\usepackage{pgf}
\usepackage{filecontents}

\def\Ours{CRT-6D}

\def\wacvPaperID{1789} 

\wacvalgorithmstrack   

\wacvfinalcopy 

\ifwacvfinal
\usepackage{hyperref}
\hypersetup{colorlinks=true,linkcolor=blue}
\else
\usepackage{hyperref}
\hypersetup{colorlinks=true,linkcolor=blue}
\fi

\begin{document}

\title{\Ours: Fast 6D Object Pose Estimation with Cascaded Refinement Transformers}

\author{Pedro Castro\\
Imperial College London\\
{\tt\small p.castro18@imperial.ac.uk}
\and
Tae-Kyun Kim\\
Imperial College London, KAIST\\
{\tt\small tk.kim@imperial.ac.uk}
}

\maketitle
\thispagestyle{empty}

\input{sections/abstract}


\input{sections/intro}
\input{sections/literature}
\input{sections/method}
\input{sections/experiments}
\input{sections/conclusion}

\pagebreak

{\small
\bibliographystyle{ieee_fullname}
\bibliography{egbib}
}

\end{document}

%% file: sections/abstract.tex
\begin{abstract}

Learning based 6D object pose estimation methods rely on computing large intermediate pose representations and/or iteratively refining an initial estimation with a slow \textit{render-compare} pipeline.
This paper introduces a novel method we call \textbf{C}ascaded Pose \textbf{R}efinement \textbf{T}ransformers, or \textbf{\Ours}. We replace the commonly used dense intermediate representation with a sparse set of features sampled from the feature pyramid we call \textbf{OSKFs}(\textbf{O}bject \textbf{S}urface \textbf{K}eypoint \textbf{F}eatures) where each element corresponds to an object keypoint. 
We employ lightweight deformable transformers and chain them together to iteratively refine proposed poses over the sampled OSKFs.
We achieve inference runtimes $2\times$ faster than the closest real-time state of the art methods while supporting up to 21 objects on a single model.
We demonstrate the effectiveness of \Ours ~by performing extensive experiments on the LM-O and YCB-V datasets. Compared to real-time methods, we achieve state of the art on LM-O and YCB-V, falling slightly behind methods with inference runtimes one order of magnitude higher. The source code is available at: \url{https://github.com/PedroCastro/CRT-6D}
\end{abstract}





%% file: sections/intro.tex
\section{Introduction}
\label{sec:Introduction}

Estimating the 6D pose of objects given an RGB image remains a challenging computer vision task yet indispensable in many real world applications from autonomous vehicle perception, robotics as well as augmented reality. This task entails the retrieval of a target object's 3D rotation and translation, relative to a camera, by overcoming difficult issues such as occlusion, illumination and symmetries. Depth information can used to great effect when available \cite{ssd6d, deepim}, while monocular methods tend to underperform due to lack of information. 

\begin{figure}[tb]
  \resizebox{\linewidth}{!}{\includegraphics[]{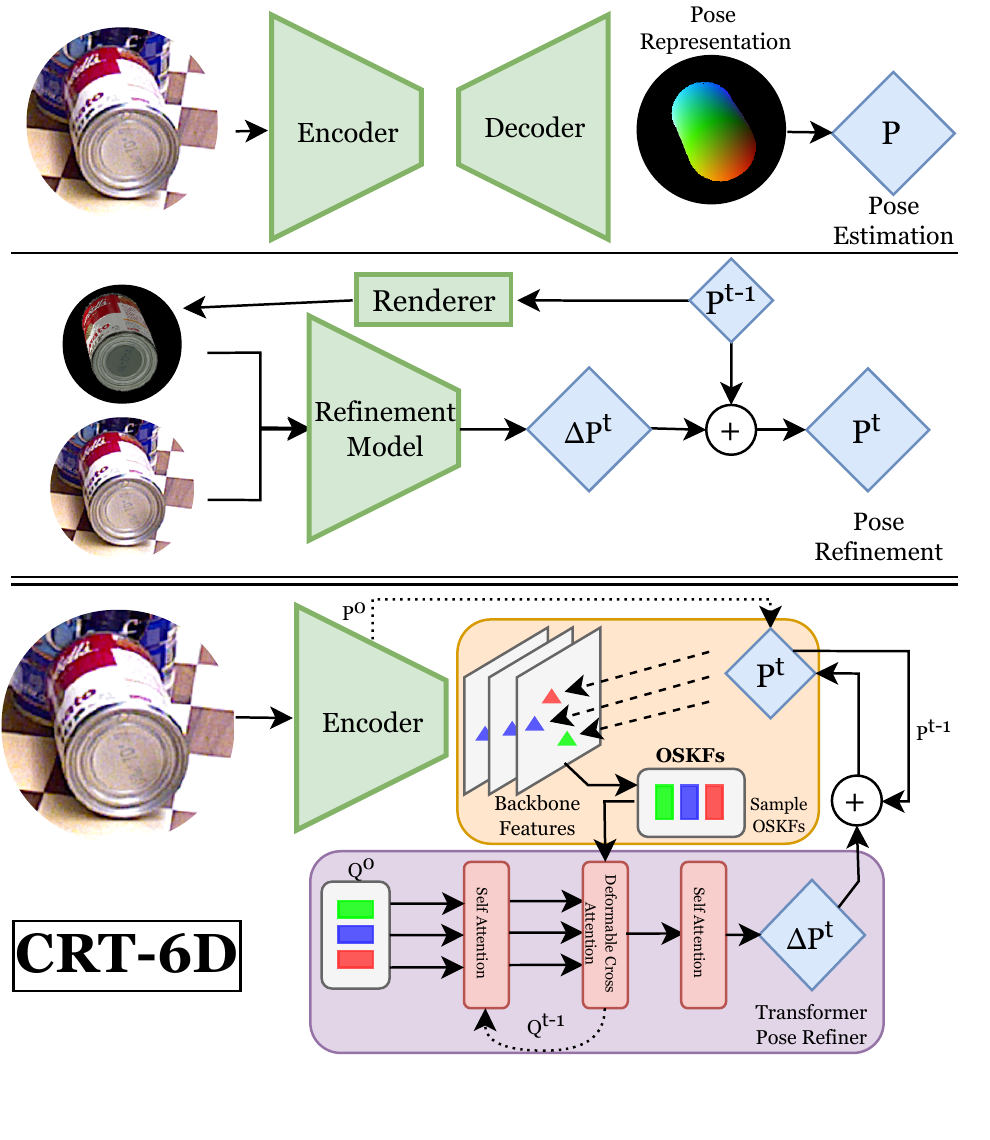}}
  \label{fig:first_page}
  \vspace{-20pt}
\caption{\textbf{Illustrative diagram of \Ours.} \Ours ~removes the decoder and pose representation from pose estimation methods, and the renderer and refinement model from the standard refinement pipelines. Instead \Ours~ replaces them with a deformable attention based refinement module achieving pose estimation and refinement within the same model. Each refinement iteration takes less than 3ms making \textbf{\Ours ~2$\times$ faster than prior real time methods and simultaneously achieving better accuracy.}}
\label{fig:init_image}
\vspace{-10pt}
\end{figure}

Recent methods utilizing Convolutional Neural Networks (CNNs) have surpassed prior classical approaches and are at the core of most recent state-of-the-art 6D object pose estimators \cite{aae,mine,pvnet,juil,bb8,gdr,sopose,cosypose,zebrapose,pix2pose,SC6D}. The computation pipeline of these methods can be roughly defined by 3 steps: 1.) The object is detected in the image (this usually done using an off-the-shelf object detector \cite{fasterrcnn,yolo}) ; 2.) Features are extracted from a cropped image, around the 2D area containing the object, using an established CNN pre-trained architecture \cite{resnet, efficientnet}; 3.) These features are transformed into an intermediate representation \cite{pvnet,pix2pose,gdr,bb8,epos} which are then used to extract pose (using PnP \cite{epnp} or other variations \cite{zebrapose,epos}) or pose is extracted directly \cite{bpnp,gdr,sopose,cosypose}. By assuming that the necessary information to extract the pose is performed by step 2.), the pose extraction step is the key to a fast and accurate estimation. Prior art has proposed several intermediate representations, e.g. NOCS, keypoint heatmaps \cite{nocs, bb8, pvnet, dpod}. These representations cover the full input crop area thereby computing it at every pixel on a significant spatial dimension, regardless of the area of the image occupied by the object, resulting in a large amount of unnecessary and expensive computations. Moreover, some require an additional slow RANSAC PnP step.
Others methods propose to directly learn the PnP operation \cite{gdr,bpnp,sopose}, and while they are shown to be faster and more precise, introduce more complexity into the model without removing the information-less regions from the computational pipeline. 

On top of these, an application might choose to refine the predicted erroneous pose. The most commonly used methods rely on a costly \textit{render-compare} iterative process \cite{deepim, dpod, cosypose,densefusion},  making them unsuitable for real-time applications. \textit{Ad-hoc} refinement methods require large models, designed and trained only for refinement and leaving the initial pose estimation as an exercise for other methods \cite{cosypose, deepim, dpod}. More recently, specifically designed approaches perform a trade-off between runtime and initialization: RePose while fast requires a great initialization \cite{repose} and SurfEmb \cite{surfemb} is very precise and robust to occlusions and symmetries but is extremely slow at inference time .

In this paper, we introduce a novel method that removes redundant computations around areas where the object is not present, while oversampling the image regions where it is. We achieve this by using a simple yet effective intermediate offset representation: \textit{Object Surface Keypoint Features} (OSKFs). 
Given an initial coarse pose, we project pre-determined object surface keypoints into the image plane.
We generate OSKFs by sampling the extracted feature pyramid at each keypoint current 2D location. Given that the intial pose is not guaranteed to be precise, we use deformable attention to guide our sampling around the original 2D location, overcoming possible errors in the coarse pose.
Therefore, we propose \textit{OSKF-PoseTransformers} (OSKF-PT), a transformer module with deformable attention mechanisms \cite{deform_detr}, where  self-attention and cross-attention operations are performed over the OSKFs set, outputting an improved pose. Since OSKFs are an inexpensive representation in terms of computation, we chain together multiple OSKF-PT in a novel Cascaded Pose Refinement (CPF) module to iteratively refine the pose in a cascaded fashion, which can be trained end-to-end. 

In summary, this paper's contributions are:

\begin{itemize}
    \item We propose Object Surface Keypoint Features (OSKF), a lightweight intermediate 6d pose offset representation, which is significantly less noisy, ignores unusable information from feature maps resulting in a more accurate pose estimation when compared to prior art and is considerably cheaper to generate than intermediate pose representations.
    \item We propose OSFK-PoseTransformer (OSFK-PT), a module that utilizes a chain of self-attention and deformable-attention layers to iteratively update an initial pose guess. Due to the lightweight nature of OSKFs, our refinement is faster than any prior refinement method, taking less than 3ms per iteration.
    \item We introduce \Ours, a fast end-to-end 6d pose estimation model, that leverages a cascaded iterative refinement over a chain of OSFK-PTs to achieve state of the art accuracy for real time 6D pose estimators on two challenging datasets, with its inference time being $100\%$ faster than the fastest prior methods.
\end{itemize}


%% file: sections/literature.tex
\section{Literature Review}
\label{sec:intro}


\noindent\textbf{Keypoint Detection.} Object pose estimation can be seen as the inverse of camera pose estimation. One can extract 6D pose by solving the PnP problem which means we can detect the pixel position of keypoints, creating the necessary 2D-3D correspondence set. Early works started by choosing the 3D bounding box of the objects as keypoints \cite{yolo6d, bb8, heatmaps}. However, the projected 3D bounding box keypoints usually lie outside the silhouette of the objects , which potentially reduces the local information extraction. This shortfall was noticed by PVNet \cite{pvnet}, which suggests the use of the surface
region to find suitable 
keypoints. 

\noindent\textbf{Dense Object Coordinate Estimation.} 
Instead of pre-selecting a few keypoints, NOCS~\cite{nocs} was proposed where for every pixel in the silhouette of the object is used to estimate the coordinate of the surface of the object (in normalized space) projected at that pixel. In other words, every point in the surface would become a keypoint and could be used for the 2D-3D correspondence set to solve PnP. Inspired by NOCS, Pix2Pose \cite{pix2pose} proposed the use of a GAN to solve issues with occlusion. DPOD \cite{dpod, epos, gdr, zebrapose} suggested using UV maps and object regions instead of a 3D coordinate system, ensuring that every point estimated lied within the object's surface.
Each of these methods has an increasingly more complex model, and while performance has been improved by each method, runtime has been overlooked.

\noindent\textbf{Direct Pose Estimation.} 
Posenet\cite{posenet} proposed learning quaternions to predict rotation on camera pose estimation tasks. In 6D object pose estimation field, PoseCNN \cite{posecnn} used Lie algebra instead. SSD-6D \cite{ssd6d} discretized the viewpoint space and learned to classify it, while using the mask to regress the distance to the camera. 
These methods are more susceptible to noise and occlusion due to their holistic approach. Moreover, these usually require an extra step to solve the ambiguity caused by egocentric orientations. Some methods learn a mapping from an intermediate representation to emulate a PnP solver, making them differentiable w.r.t. the final pose \cite{gdr, sopose, singlestage, bpnp}. This step can be used to reduce the need for symmetry hacks on dense methods. 

\noindent\textbf{Pose Estimation Refinement}
If depth is available, then Iterative Closest Point (ICP) is the most commonly used algorithm \cite{icp}. The ICP algorithm finds the correspondence between points by iteratively refining the pose that takes to align them. However, it is heavily dependent on initial pose and might converge in a local minimum. Recent learned approaches mostly rely a \textit{render-compare} pipeline, with slight variations among these methods \cite{deepim,dpod,cosypose}. More recently, Repose \cite{repose} introduced a fast iterative refinement algorithm however it requires a great initialization.

\noindent\textbf{Transformers in Pose Estimation.} Given the rising effectiveness of transformers in computer vision tasks, there have been attempts to use transformers to improve human~\cite{human_transformer1, human_transformer2}, hand~\cite{hand_transformer1} and object~\cite{DProST, yolopose, 6dvit, osop, handobjecttransformer} pose estimation. For object pose, such approaches are aimed at improving results~\cite{DProST, yolopose}, category level estimation~\cite{6dvit, osop} or hand-object interaction~\cite{handobjecttransformer}. However, these improvements come at the cost of runtime, making them unsuitable for real-time applications. Our novel approach, not only improves results, but also decreases the runtime when compared to prior object pose estimation methods.

%% file: sections/method.tex
\section{Methodology}
\label{sec:Methodology}


In this section, we detail each step of \textbf{\Ours}, our novel 6D pose estimation method. Given an image $\mathcal{I}$, the goal of \Ours ~is to predict the $P_i = [R_i|t_i]$, the 6D pose of the objects targeted by the camera, where $i$ refers to the $i^{th}$ object in the set of N objects $\mathcal{O} = \{ \mathcal{O}_i~|~i = 0, · · · , N{-}1 \}$ present in the image. We follow the setup of other methods \cite{gdr,sopose,3drcnn} and  disentangle pose from object detection which means we use an off-the-shelf detector to crop out regions of the image where objects are present.
These regions are then independently processed and fed to \Ours~ for pose estimation.

\begin{figure}[tb]
  \resizebox{\linewidth}{!}{\includegraphics[]{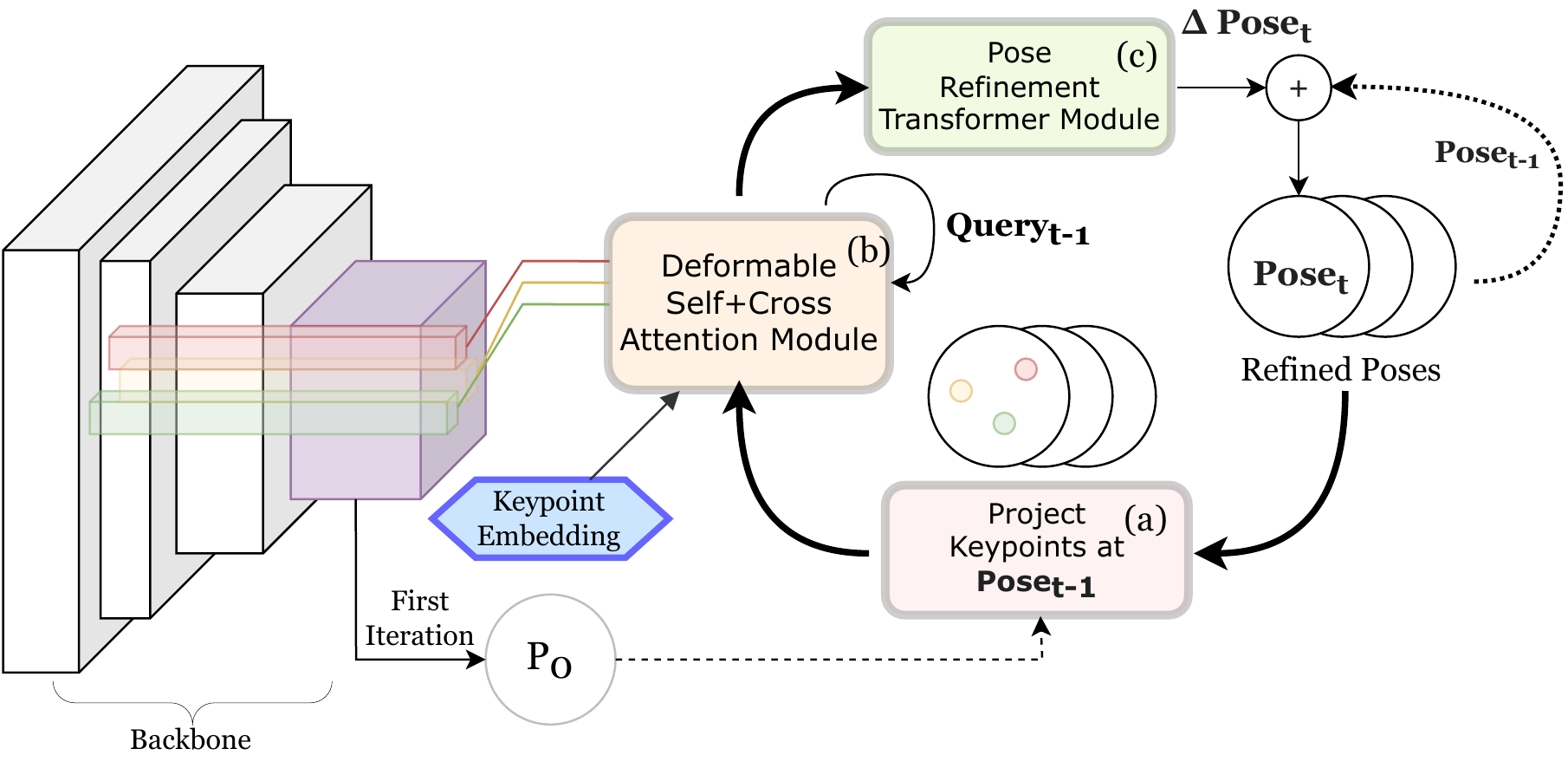}}
  \label{fig:crt6d}
  \vspace{-5pt}
\caption{\textbf{Illustration of \textbf{\Ours}.} We start by generating our feature pyramid. Using the last feature level, we generate a pose estimation $P^0$, which will serve as the initial reference for the refinement module. \textbf{(a)} Using $P^0$, we project a set of surface keypoints $\mathcal{S}$  into the image plane. \textbf{(b)} Our deformable attention mechanism uses the location of the set of projected keypoints to sample the feature pyramid. \textbf{(c)} We then perform self-attention over the sampled feature set and estimates an offset pose $\Delta P$ such that $P^t = \hat{P}^{t-1} + \Delta{P^t}$, with $t$ being the refinement step.}
\label{fig:init_image}
\vspace{-15pt}
\end{figure}

\subsection{Coarse Pose Estimation} \label{sec:coarse}

We use a Resnet34~\cite{resnet} as the backbone for \Ours. While other backbones could have been used, Resnets present fairer comparisons with prior art as they are used by most 6D pose approaches\cite{gdr,sopose,zebrapose,repose,cdpn}. We use the backbone to build a multi-scale feature pyramid $\mathcal{F} = \{ \mathcal{F}_l~|~l = 1, · · · , L \}$ of sizes $[s/4,s/8, s/16, s/32]$, with $L=4$s.

\Ours ~starts by estimating a coarse pose $P^0_{6D}=[R^0|t^0]$ using a simple MLP: $FC_{\theta}(F_3) = R^0_{6D}, \Tilde{t}^0$, where $\theta$ are learnable parameters. 
To recover the rotation matrix $R = [R_1,R_2,R_3]$, we use the 6D rotation representation $R_{6D} = [r_1, r_2]$ introduced in \cite{continuous} and used in prior methods \cite{gdr,sopose,cosypose} to great success:

\begin{equation}
    \begin{cases}
    R_1 = r_1 \\
    R_3 = R_1 \times r_2 \\
    R_2 = R_3 \times R_2
    \end{cases}
    ,
\end{equation}

where $r1$ and $r2$ are unit vectors. Due to the projective function applied by the camera, the appearance of an object is affected not only by its orientation but also by its position. Since we are working with a cropped image and camera intrinsics $K$ are known, predicting $P$ becomes a one-to-many function, where the same appearance might correspond to different egocentric orientations as pointed out by 3D-RCNN~\cite{3drcnn}. A common strategy to solve this problem, which we also adopt, is to estimate the allocentric orientation and apply a transformation to recover the egocentric orientation at inference time.

Wang \etal~\cite{gdr} showed that choosing an adequate translation representation has an effect on the performance of the method. While our goal is the global translation $t=[t_x, t_y, t_z]$, this information cannot be directly recovered by \Ours~ due to the cropping step. Therefore we use $t'=[O_x, O_y, t_z]$ which can be used to recover $t$ via back-projection. \Ours~adopts the scale-invariant representation $\Tilde{t} = [\gamma_x, \gamma_y, \gamma_z]$ \cite{gdr, sopose}:

\begin{equation}
    \begin{cases}
    \gamma_x=(O_x - c_x)/s_{bbox} \\
    \gamma_y=(O_y - c_y)/s_{bbox} \\
    \gamma_z=t_z/r_{bbox}
    \end{cases}
\end{equation}

where the scale of the crop bounding box $s_{bbox}=max(w_{bbox}, h_{bbox})$ and ratio $r=s_{bbox}/s$, with $s$ referring to the original size of the image.

\subsection{\textbf{O}bject \textbf{S}urface \textbf{K}eypoint \textbf{F}eatures - OSKF}

Refinement methods usually rely on a prior independent powerful pose estimator \cite{cosypose,repose,deepim} slowing both inference and training. In contrast, \Ours~ is designed to estimate and refine by reusing the multi-scale feature pyramid. 
We replace the rendering step used by pose refiners \cite{cosypose,deepim,repose} and instead of using a pose representation, we generate a pose offset representation $\mathcal{Z}$, a compact set of backbone features sampled at 2D keypoint locations $\mathcal{P}^t = \{\pi(\mathcal{S}_k^t, P_{6D}^t, K_{cam})~|~k = 1,..., K \}$, where $\pi$ is the projection function, $K_{cam}$ are the known camera intrinsic parameters, $P_{6D}^t$ is the reference pose, $\mathcal{S}$ is a set of pre-defined keypoints, with $t$ denotes the iteration step. Since the input image $I$ suffers no perturbations from the pose estimation step to the refinement step, or even between refinement iterations, we recycle these features instead of recomputing them at every iteration, in contrast to prior methods \cite{dpod, deepim, cosypose, repose}. 

We propose OSKFs $\mathcal{Z} = \{\mathcal{Z}_k~|~k = 1,..., K \}$, a lightweight offset pose representation generated by sampling the feature pyramid at all spatial scales: 

\begin{equation}
    \mathcal{Z} = \{\mathcal{F}_l(\mathcal{P}_k)~|~k = 1,..., K, ~l = 1,..., L \} 
\end{equation}
where $l$ denotes the \textit{$l^{th}$} element of the feature pyramid.

The set of object keypoints $\mathcal{S} = \{ \mathcal{S}_k~|~k = 1,..., K \}$ chosen for OSKFs are generated using the farthest point sampling algorithm \cite{pvnet}, where $K$ is a hyperparameter of the number of used keypoints.
Each of these features represent the local information around $\mathcal{P}$, with coarse information at higher feature levels and finer on lower. The following modules will learn to recover the pose offset embedded in $\mathcal{Z}$.

\begin{figure}[tb]
  \resizebox{\linewidth}{!}{\includegraphics[]{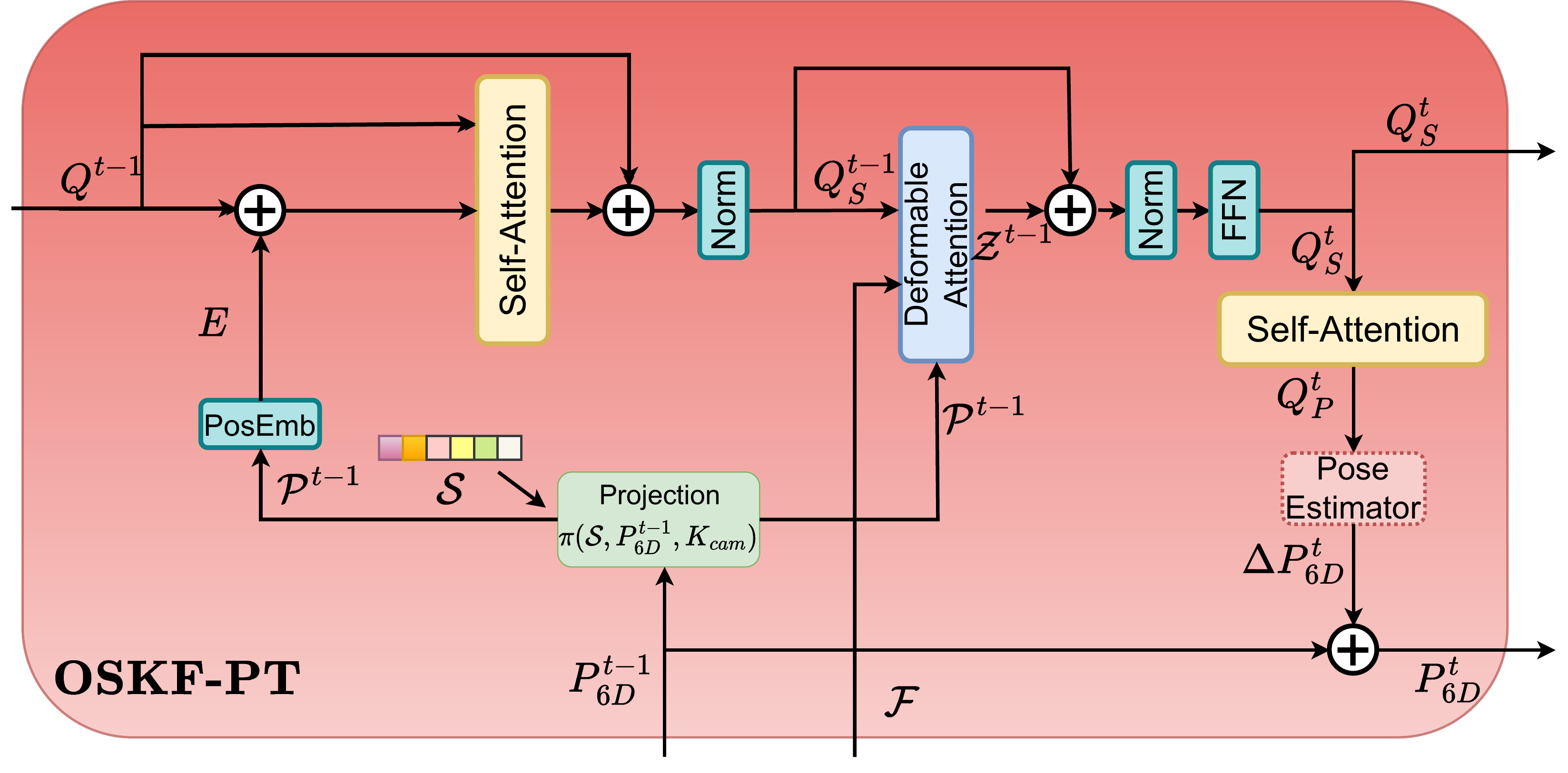}}
  \label{fig:transformer}
  \vspace{-15pt}
\caption{\textbf{Illustration of \textbf{OSKF-PT}.} OSKF-PT receives as an input the previous $Q^{t-1}$, $P^{t-1}_{6D}$ and $\mathcal{F}$, the last one being the same on every iteration. It outputs a new refined pose $P^{t}_{6D}$ and $Q^{t}$, an initial query to guide the next iteration.}
\label{fig:transformer}
\vspace{-15pt}
\end{figure}

\subsection{OSKF Pose Transformer (OSKF-PT)}

The untopologized set of OSKFs $\mathcal{Z}$ presents a use case for transformers which are able to process data without introducing implicit spatial bias \cite{attention} allowing the attention mechanisms to learn spatial and  structural relationships. \Ours~ is designed to interpret $\mathcal{Z}$ and learn to output the offset pose $\Delta P$ such that:
\begin{equation} \label{eq:offset}
    P^t_{6D} = \hat{P}_{6D}^{t-1} + \Delta{P}^t_{6D} ,
\end{equation} 
at the $t$ refinement step.

However, the estimated pose is not guaranteed to be correct nor the keypoints unoccluded. For this reason, the optimal information for a keypoint might not lie close to the 2D position of that object.
Inspired by Deformable-DETR \cite{deform_detr} we introduce OFKF-PT where we make use of deformable attention mechanisms to learn to sample information around the reference keypoint position. A diagram of the pose transformer module OFKF-PT is showed in Fig \ref{fig:transformer}.

OFKF-PT are composed by a self-attention layer, a deformable self attention layer and another attention layer for pose estimation, in this order.
We start by performing self-attention over a query matrix $Q$, where each row corresponds to a keypoint in the set $\mathcal{S}$. The multi-head attention operation takes $\hat{Q}$, $\hat{Q}$ and $Q$ as query, key and value, where $\hat{Q} = Q + E$. $E_k = \textit{PosEmb}(\mathcal{P}_k)$ are the positional embeddings for keypoint $k$ and \textit{PosEmb} is the high frequency cosine embedding for position parameterization \cite{attention}. Following \cite{deform_detr}, we then apply residual addition and layer normalization. We denote the output of this operation as $Q_S$.

The multi-scale deformable attended OSKFs are formally defined by: 

\begin{equation}\label{eq:attention_deform}
    \Tilde{\mathcal{Z}}_{kl} =\sum_{j=0}^{J} A_{ljk} W \mathcal{F}_l(\mathcal{P}_k + \Delta\mathcal{P}_{ljk})
\end{equation}
where $\Delta\mathcal{P}_{ljs}$ refers to the predicted sampling offset for the $j^\textit{th}$ deformable position, spatial level $l$ and object keypoint $k$. $J$ refers to the number of sampling points used for deformation.
$\Delta\mathcal{P}_ljs$ and $A_{ljs}$ are computed by a simple linear transformations of $Q_S$. 

We apply a self-attention operation over $\Tilde{\mathcal{Z}}$ in order to capture long distance information which we denote as $Q_P$.
We chose to use a global pooling operation over $Q_P$ instead of using a CLS-Token (which is shown to make no difference in practice \cite{16x16words}), followed by a small MLP to output $[\Delta R , \Delta\gamma_x, \Delta\gamma_y, \Delta\gamma_z ]^t $. We update the reference pose $P^t$:

\begin{equation}
    \begin{cases}
        R^t = \Delta R^t~R^{t-1}\\
        \gamma_{\{x,y\}}^t = \gamma_{\{x,y\}}^{t-1} + \Delta\gamma_{\{x,y\}}^t\\
        \gamma_{z}^t = \gamma_{z}^{t-1} \cdot (1 + tanh(\Delta\gamma_{z}^t))
    \end{cases} .
\end{equation}

We found that using a \textit{tanh} parameterization over $\Delta\gamma_{z}$ performed better than linear \cite{cosypose} or exponential \cite{deepim}. 

For our experiments, we chain together 3 OFKF-PT.
The output of the deformable attention operation is reused on the next iteration, as the input matrix becomes $Q^t = \Tilde{\mathcal{Z}}^{t-1}$. The initial query $Q^0$ is a learnable embedding.

\subsection{Objective Functions}

The main goal of \Ours ~is to process an image and produce the 6D pose of the target object. To perform this operation, it goes through an iterative refinement process. At each iteration, we compute a valid 6d pose and reuse it on the next iteration. Due to the simplified objective of \Ours, which does not output an intermediate pose representation, our overall objective $\mathcal{L}$ is guided by the pose estimation error and can be defined as:

\begin{equation}
    \mathcal{L} = \lambda\mathcal{L}^0 + (1-\lambda)\sum^{N}_{i=1}\mathcal{L}^i
\end{equation}

with $i$ indicates the refinement iteration, $N=3$ is fixed for all experiments and $\mathcal{L}^0$ the loss for the coarse estimation. 
The pose loss $\mathcal{L}$ is disentangled and separated into rotation and position loss: 

\begin{equation}
    \mathcal{L} = \alpha\mathcal{L}_R + \mathcal{L}_{pos} .
\end{equation}

\begin{table*}[]
\resizebox{\textwidth}{!}{%
\begin{tabular}{c|cccc|ccccc}
\Xhline{5\arrayrulewidth}    
Type & \multicolumn{5}{c|}{Pose Estimation}                                                                                                                    & \multicolumn{2}{c|}{Refinement}                                   & Hybrid    \\ \hline 
N.S.O.                        & \multicolumn{1}{c|}{1}     & \multicolumn{1}{c|}{8}    & \multicolumn{1}{c|}{1}    & \multicolumn{1}{c|}{8}       & \multicolumn{1}{c|}{1}             & \multicolumn{1}{c|}{1}      & \multicolumn{1}{c|}{8}             & 8    \\ \hline
Method                   & \multicolumn{1}{c|}{PVNet~\cite{pvnet}} & \multicolumn{1}{c|}{GDR~\cite{gdr}}  & \multicolumn{1}{c|}{GDR~\cite{gdr}}  & \multicolumn{1}{c|}{SO-Pose~\cite{sopose}} & \multicolumn{1}{c|}{ZebraPose~\cite{zebrapose}}     & \multicolumn{1}{c|}{RePose~\cite{repose}} & \multicolumn{1}{c|}{DeepIM~\cite{deepim}}        & \textbf{\Ours} \\ \hline\hline
Ape                     & \multicolumn{1}{c|}{15.8}  & \multicolumn{1}{c|}{44.9} & \multicolumn{1}{c|}{46.8} & \multicolumn{1}{c|}{48.4}    & \multicolumn{1}{c|}{\underline{55.2}}         & \multicolumn{1}{c|}{31.1}   & \multicolumn{1}{c|}{\textbf{59.2}} & 53.4 \\ \hline
Can                     & \multicolumn{1}{c|}{63.3}  & \multicolumn{1}{c|}{79.7} & \multicolumn{1}{c|}{90.8} & \multicolumn{1}{c|}{85.8}    & \multicolumn{1}{c|}{\textbf{94.9}} & \multicolumn{1}{c|}{80.0}   & \multicolumn{1}{c|}{63.5}          & \underline{92.0} \\ \hline
Cat                     & \multicolumn{1}{c|}{16.7}  & \multicolumn{1}{c|}{30.6} & \multicolumn{1}{c|}{40.5} & \multicolumn{1}{c|}{32.7}    & \multicolumn{1}{c|}{\textbf{56.6}} & \multicolumn{1}{c|}{25.6}   & \multicolumn{1}{c|}{26.2}          & \underline{42.0} \\ \hline
Driller                 & \multicolumn{1}{c|}{65.7}  & \multicolumn{1}{c|}{67.8} & \multicolumn{1}{c|}{82.6} & \multicolumn{1}{c|}{77.4}    & \multicolumn{1}{c|}{\textbf{94.7}} & \multicolumn{1}{c|}{73.1}   & \multicolumn{1}{c|}{55.6}          & \underline{81.4} \\ \hline
Duck                    & \multicolumn{1}{c|}{25.2}  & \multicolumn{1}{c|}{40.0} & \multicolumn{1}{c|}{46.9} & \multicolumn{1}{c|}{48.9}    & \multicolumn{1}{c|}{\textbf{60.9}} & \multicolumn{1}{c|}{43.0}   & \multicolumn{1}{c|}{\underline{52.4}}          & 44.9 \\ \hline
Eggbox*                 & \multicolumn{1}{c|}{50.2}  & \multicolumn{1}{c|}{49.8} & \multicolumn{1}{c|}{54.2} & \multicolumn{1}{c|}{52.4}    & \multicolumn{1}{c|}{\textbf{64.7}} & \multicolumn{1}{c|}{51.7}   & \multicolumn{1}{c|}{\underline{63.0}}          & 62.7 \\ \hline
Glue*                   & \multicolumn{1}{c|}{49.6}  & \multicolumn{1}{c|}{73.7} & \multicolumn{1}{c|}{75.8} & \multicolumn{1}{c|}{78.3}    & \multicolumn{1}{c|}{\textbf{84.5}} & \multicolumn{1}{c|}{54.3}   & \multicolumn{1}{c|}{71.7}          & \underline{80.2} \\ \hline
Holepuncher             & \multicolumn{1}{c|}{36.1}  & \multicolumn{1}{c|}{62.7} & \multicolumn{1}{c|}{60.1} & \multicolumn{1}{c|}{75.3}    & \multicolumn{1}{c|}{\textbf{83.2}} & \multicolumn{1}{c|}{53.6}   & \multicolumn{1}{c|}{52.5}          & \underline{74.3} \\ \hline \hline
Average                 & \multicolumn{1}{c|}{40.8}                       & \multicolumn{1}{c|}{56.1}                      & \multicolumn{1}{c|}{62.2}                      & \multicolumn{1}{c|}{62.3}    & \multicolumn{1}{c|}{\textbf{74.3}} & \multicolumn{1}{c|}{51.6}   & \multicolumn{1}{c|}{55.5}          & \underline{66.3} \\
\Xhline{5\arrayrulewidth}             
\end{tabular}
}
\caption{\textbf{Comparison study on LM-O.} We present the results for ADD(-S) metric and compare them to state of the art. We are outperformed only by ZebraPose~\cite{zebrapose}, a method with an inference time of $\sim 191$ms for a single object, while \textbf{\Ours}~ estimates the pose for all objects in a single LM-O image ($\sim 8$ objects) in 36ms.
Best results are bolded while second best are underlined. \textit{N.S.O.} refers to the number of objects supported by a model and $^*$ denotes symmetric objects.}
\label{tab:lmo}
\end{table*}

Recalling the pose parameterization described in Sec. \ref{sec:coarse}, the loss functions are defined as:

\begin{equation}
    \begin{cases}
        \mathcal{L}_R = \underset{x\in \mathcal{P}}{avg} ||Rx - \hat{R}x||_1\\
        \mathcal{L}_{pos} = ||\gamma_x - \hat{\gamma_x} , \gamma_y - \hat{\gamma_y} , \gamma_z - \hat{\gamma_z}||_1 
    \end{cases} ,
\end{equation}

where $\hat{\cdot}$ refers to the groundtruth data. When the target object is symmetric, a variation of $\mathcal{L}_R$ is used \cite{nocs}, while $\mathcal{L}_{pos}$ is invariant to symmetries. Note that while our refinement module outputs an offset pose $\Delta P$, the transformation in Eq.~\ref{eq:offset} is differentiable which means \Ours~ can 
be directly optimized through the set of predicted poses $P_{6D} = \{P^t_{6D} | t=0,...,N\}$

%% file: sections/experiments.tex
\section{Experiments}
\label{sec:Introduction}

We conducted experiments on two benchmark dataset LM-O\cite{linemod} and YCB-V\cite{ycb} where we present strong evidence of our method's potential. We also show through ablation studies our key contributions, including the improvements stemming from the use of iterative refinement and the high accuracy it achieves with an impressively low inference time. Results for all BOP datasets~\cite{bop} are available on the \href{https://bop.felk.cvut.cz/method_info/276/}{challenge website}.

\subsection{Experimental Setup}

\noindent \textbf{Datasets Setup.} The commonly used Linemod dataset (LM) \cite{linemod} has become sasturated with most recent methods achieving over 95\% accuracy \cite{cosypose, gdr, sopose}. For this reason we adopt our experiments on the more challenging Linemod Occlusion (LM-O), a subset of 1214 LM images, where $\sim~8$ objects are annotated on every image. For LM-O, in accordance with prior art \cite{gdr,sopose,zebrapose}, we make use of the available LM real images, where $\sim~1200$ images are available per object. 
We also present experiments on YCB-V, a larger dataset with 21 target objects, some with very challenging symmetries. For this dataset over 100k real images are available for training. However, the dataset is generated through video resulting in similar frames where the objects are seldom fully visible.

On top of real images, we also make use of synthetic data. For a fair comparison, we employ the readily available PBR splits \cite{pbr}, available for both LM-O and YCB-V, a dataset of photo and physically realistic synthetic images containing the target models with challenging poses and under heavy occlusion.
We also perform common on-the-fly image augmentations such as color jittering, blur and noise as well as more complex operations such as in-plane image rotations and background removal. 
For experiments under the BOP~\cite{bop} setup, LM-O methods are trained only with PBR synthetic data. 
We also implement the Dynamic Zoom-In (DZI) \cite{cdpn} in order to be robust to detection errors. During training we apply uniform perturbations to the center and scale of the bounding box. At test time, we found that increasing the detection bounding boxes by 20\%, to ensure the object is fully visible, yielded the best results.

\noindent\textbf{Implementation details.}
We implement \Ours~ using PyTorch~\cite{pytorch}. We use 8 heads and 4 points for deformable attentions as suggested in Deformable-DETR \cite{deform_detr}. The model is trained in an end-to-end fashion, including the cascaded refinement step. All ablation experiments are optimized with same number of training iterations. For LM-O experiments, \Ours~ is optimized for 250k iterations with batch size of 32, with PBR images composing $50\%$ of the batch, or $100\%$ if under BOP standards \cite{bop}. YCB-V is trained for 350k, with the same PBR ratio and batch size. We use the Ranger optimizer \cite{ranger} starting at a learning rate $10^{-4}$ with a cosine annealing schedule starting at 85\% of training. Unlike similar methods \cite{detr, deform_detr,landmarks}, we found in early experiments that choosing a lower learning rate for the backbone weights was not ideal.
For the first 20\% of the iterations we set $\lambda=0$ because $P^0$ starts with very poor pose estimations, which does not allow \Ours~ to learn. For the rest of training $\lambda=\frac{N-1}{N}$ where $N$ is the number of pose refiners used. We set $\alpha=3$ for all experiments.

\begin{figure}[b]
  \vspace{-10pt}
    \resizebox{\columnwidth}{!}{%
      \input{resources/figure.pgf}
      }
  \vspace{-10pt}
\caption{\textbf{Visualization of the inference time difference between state of the art methods.} \Ours ~is faster, even after 3 refinement steps, than prior approaches and has very competitive results compared to methods 10$\times$ slower. We measure the average time it takes to estimate the pose for \underline{all objects in an image} ($\sim$4.75 per image on YCB-V). 
ZebraPose \cite{zebrapose} and SurfEmb~\cite{surfemb} are omitted as their results would lie outside the runtime range, with estimation taking over 250ms and 2000ms \underline{per crop}, respectively.
}
\label{fig:runtime_ycbv}
  \vspace{-5pt}
\end{figure}
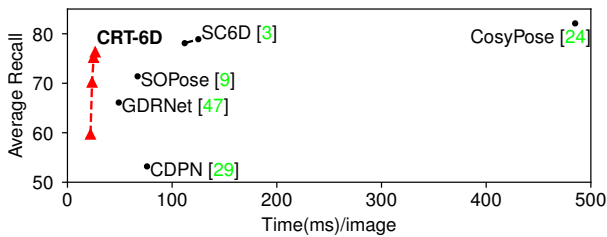

During testing, we use the same detections per dataset for all experiments. For LM-O we use the publicly available 2D detections from Faster-RCNN  utilized by \cite{gdr, sopose, zebrapose}. For YCB-V, we also use publicly available detections from FCOS\cite{fcos} trained by \cite{cdpn}. For BOP results, we the standardized detections provided by the challenge. 

\noindent \textbf{Evaluation Metrics.}
For a fair comparison, we emulate the same practice as prior methods. For LM-O, we present our results over the ADD(-S) metric \cite{bb8,bop}. Under the ADD(-S) metric, a pose is considered correct if the distance of the transformed model points from the groundtruth points is below $d\cdot10\%$, $d$ being the object's diameter. For symmetric objects, we take the distance from each transformed point to the closest groundtruth and apply the same threshold \cite{bop, linemod}. When experimenting on the YCB-V dataset, we also take the Area Under the Curve (AUC) of the ADD metric, with a maximum threshold distance of 10cm \cite{posecnn}. For more detailed experiments we also measure accuracy in terms of $n^{\circ}, n~$cm which considers a pose valid if both the rotation and translation fall under the defined thresholds.
Under BOP standards, we present the average recalls used by the challenge: $AR_{vsd}$, $AR_{mssd}$ and $AR_{mspd}$, along with their mean. We refer the reader to the BOP challenge \cite{bop} for more information about these metrics. The inference time measurements, for both \Ours~ and prior methods, were all made using publicly available code on a GTX1080ti. For simplicity, we ignore detection time as most methods are evaluated using the same detections.

\begin{table*}[]
\resizebox{\textwidth}{!}{%
\begin{tabular}{c|cccc|cccc|c}
\Xhline{5\arrayrulewidth}
\multirow{2}{*}{Method} & \multicolumn{4}{c|}{LM-O}                                                                                                        & \multicolumn{4}{c|}{YCB-V}                                                                                                       & \multirow{2}{*}{\textbf{Mean AR}} \\ \cline{2-9}
                        & \multicolumn{1}{c|}{$AR_{VSD}$}          & \multicolumn{1}{c|}{$AR_{MSSD}$}         & \multicolumn{1}{c|}{$AR_{MSPD}$}         & \textbf{AR}             & \multicolumn{1}{c|}{$AR_{VSD}$}          & \multicolumn{1}{c|}{$AR_{MSSD}$}         & \multicolumn{1}{c|}{$AR_{MSPD}$}         & \textbf{AR}             &                                            \\ \hline
EPOS~\cite{epos}                      & \multicolumn{1}{c|}{0.389}          & \multicolumn{1}{c|}{0.501}          & \multicolumn{1}{c|}{0.750}          & 0.547          & \multicolumn{1}{c|}{0.626}          & \multicolumn{1}{c|}{0.677}          & \multicolumn{1}{c|}{0.783}          & 0.695          & 0.621                                          \\ \hline
GDR-Net~\cite{gdr}                   & \multicolumn{1}{c|}{-}              & \multicolumn{1}{c|}{-}              & \multicolumn{1}{c|}{-}              &       \multicolumn{1}{c|}{-}      & \multicolumn{1}{c|}{0.584}          & \multicolumn{1}{c|}{0.674}          & \multicolumn{1}{c|}{0.726}          & 0.661          & -                                            \\ \hline
SO-Pose~\cite{sopose}                   & \multicolumn{1}{c|}{0.442}          & \multicolumn{1}{c|}{0.581}          & \multicolumn{1}{c|}{{\underline{0.817}}}    &     0.613     & \multicolumn{1}{c|}{0.652}          & \multicolumn{1}{c|}{0.731}          & \multicolumn{1}{c|}{{0.763}}    & \multicolumn{1}{c|}{0.715}           & 0.664                                      \\ \hline
SurfEmb~\cite{surfemb}                  & \multicolumn{1}{c|}{-}               & \multicolumn{1}{c|}{-}              & \multicolumn{1}{c|}{-}              & {\underline{0.656}}    & \multicolumn{1}{c|}{-}              & \multicolumn{1}{c|}{-}              & \multicolumn{1}{c|}{-}              & 0.718          & 0.687                                     \\ \hline
CosyPose~\cite{cosypose}                  & \multicolumn{1}{c|}{{\underline{0.480}}}    & \multicolumn{1}{c|}{{\underline{0.606}}}    & \multicolumn{1}{c|}{0.812}          & 0.633          & \multicolumn{1}{c|}{\textbf{0.772}} & \multicolumn{1}{c|}{\textbf{0.842}} & \multicolumn{1}{c|}{\textbf{0.850}} & \textbf{0.821} & \textbf{0.727}                                 \\ \hline
SC6D~\cite{SC6D}                 & \multicolumn{1}{c|}{-}              & \multicolumn{1}{c|}{-}              & \multicolumn{1}{c|}{-}              &          -      & \multicolumn{1}{c|}{0.695}          & \multicolumn{1}{c|}{\underline{0.796}}          & \multicolumn{1}{c|}{\underline{0.804}}          &    \underline{0.765}       & -                                            \\ \hline
\textbf{\Ours}                    & \multicolumn{1}{c|}{\textbf{0.504}} & \multicolumn{1}{c|}{\textbf{0.640}} & \multicolumn{1}{c|}{\textbf{0.837}} & \textbf{0.660} & \multicolumn{1}{c|}{{\underline{0.707}}}    & \multicolumn{1}{c|}{{0.776}}    & \multicolumn{1}{c|}{0.774}          & {0.752}    & {\underline{0.706}}          \\      
\Xhline{5\arrayrulewidth}             
\end{tabular}
}
\vspace{1pt}
\caption{\textbf{LM-O and YCB-V datasets under the BOP standards~\cite{bop}.} We present the results for all metrics used on the challenge with \textit{Mean AR} referring to average \textit{AR} over both datasets. 
Best results are \textbf{bolded} while second best are \underline{underlined}.}
\vspace{-5pt}
\end{table*}

\subsection{Runtime comparison to state of the art.}

DeepIM~\cite{deepim} is a landmark work on learned pose refinement. It is however very slow, inference time being 41ms per object per iteration (two are recommended), without taking into account the model used for initialization. More recently CosyPose~\cite{cosypose} performs a two-in-one model, where two CNN are employed and fully trained, one performs an initial coarse estimation while the second one refines it. However, due to the massive size of the models used, their inference time is $\sim 100$ms for a single object.  Repose~\cite{repose} proposed a faster refinement method at 18ms with 5 iterations however they require a good initialization (they use PVNet \cite{pvnet} which itself takes over 25ms) and it only support a single object per model.

In contrast, \Ours~takes on average 26ms and 34ms for YCB-V and LM-O images (each with $\sim 4$ and $\sim 6$ object instances) for all objects on a single model, making our method one order of magnitude faster than other refinement methods. Recent state of the art real time methods, GDR-Net~\cite{gdr}and SO-Pose~\cite{sopose} take $2\times$ and $3\times$ longer, respectively. Moreover, \Ours~not only is faster but more accurate than both these methods, on both LM-O and YCB-V.
A more detail comparison to other methods w.r.t. inference time can be seen in Fig. \ref{fig:runtime_ycbv}.

\begin{figure}[t]
\vspace{-5pt}
    \resizebox{\columnwidth}{!}{%
    \centering 
    \begin{subfigure}{\textwidth}
      \fbox{\includegraphics[width=\linewidth]{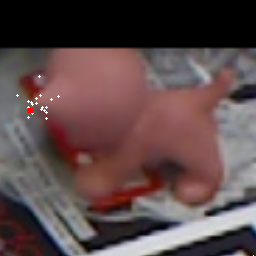}}
      \label{}
    \end{subfigure}\hfil 
    \begin{subfigure}{\textwidth}
      \fbox{\includegraphics[width=\linewidth]{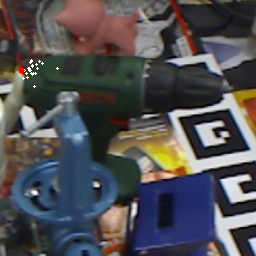}}
      \label{}
    \end{subfigure}\hfil 
    \begin{subfigure}{\textwidth}
      \fbox{\includegraphics[width=\linewidth]{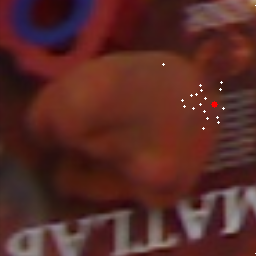}}
      \label{}
    \end{subfigure}
    }
    \vspace{-6pt}
    \medskip 
    \resizebox{\columnwidth}{!}{%
    \begin{subfigure}{\textwidth}
      \fbox{\includegraphics[width=\linewidth]{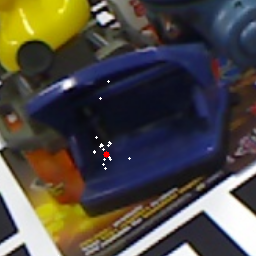}}
      \label{fig:4}
    \end{subfigure}\hfil 
    \begin{subfigure}{\textwidth}
      \fbox{\includegraphics[width=\linewidth]{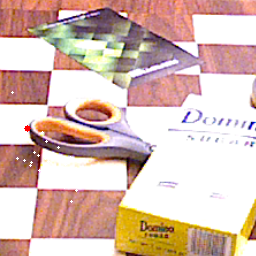}}
      \label{fig:5}
    \end{subfigure}\hfil 
    \begin{subfigure}{\textwidth}
      \fbox{\includegraphics[width=\linewidth]{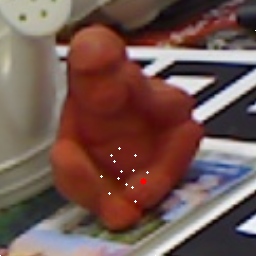}}
      \label{fig:6}
    \end{subfigure}
    }
    \medskip 
    \resizebox{\columnwidth}{!}{%
    \begin{subfigure}{\textwidth}
      \fbox{\includegraphics[width=\linewidth]{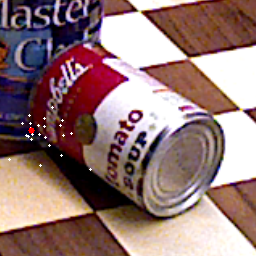}}
      \label{fig:4}
    \end{subfigure}\hfil 
    \begin{subfigure}{\textwidth}
      \fbox{\includegraphics[width=\linewidth]{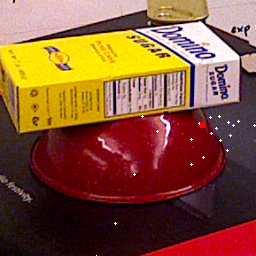}}
      \label{fig:5}
    \end{subfigure}\hfil 
    \begin{subfigure}{\textwidth}
      \fbox{\includegraphics[width=\linewidth]{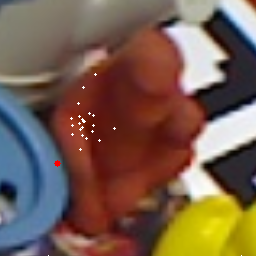}}
      \label{fig:6}
    \end{subfigure}
    }
\vspace{-5pt}
\caption{\textbf{Deformable attention sampling locations.} The red circle indicates the reference point while the white circles indicate deformed sample positions. On the last column, we show attention sampling for the same object and keypoint, with the first row having the keypoint visible whereas on the bottom two the keypoint is self or externally occluded. When the keypoint is occluded, attention learns to sample clues elsewhere on the object silhouette.}
\label{fig:attention}
\vspace{-15pt}
\end{figure}

\subsection{Accuracy comparison to the State of the Art}

\noindent\textbf{LM-O results.} 
We present the results under the ADD(-S) metric for LM-O on Tab. \ref{tab:lmo}. We present competitive results, with the second best overall accuracy behind ZebraPose \cite{zebrapose}. 
Compared with real-time methods, \Ours~achieves a $6.4\%$ improvement over SO-Pose~\cite{sopose}, the former state of the art. 
Under the BOP challenge rules, we achieve state of the art performance, reaching an Average Recall of $67.2\%$, beating all other RGB based methods, regardless of their inference time. We achieve this result because our method is extremely robust to occlusion due to our use of deformable attention and the fact that it can \textit{attent} to regions far from the reference position when the the keypoint is occluded (see Fig. \ref{fig:attention}).

\noindent\textbf{YCB-V results.} On YCB-V, we present extremely competitive results by achieving $72.1\%$ and and a state of the art $87.5$ ADD-(S) and AUC of ADD(-S), respectively. Compared to closest real-time methods, we outperform GDR-Net\cite{gdr} and SO-Pose\cite{sopose} by $46\%$ and $27\%$ on the ADD(-S) metric, $9\%$ and $4\%$ on AUC of ADD(-S). When compared to slower methods, our method is surpassed on the ADD(-S) only by ZebraPose~\cite{zebrapose}, which is an order of magnitude slower than our method. We also present YCB-V results under the BOP standard, where we are only outperformed by a slower method. 

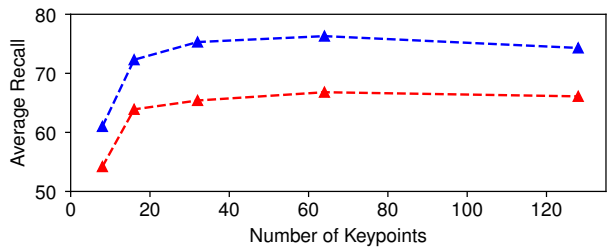
\begin{figure}[b]
    \resizebox{\columnwidth}{!}{%
      \input{resources/keypoints.pgf}
      }
\caption{\textbf{Experiments over number of OSKFs used.} With this ablation, we show that \Ours~is not reliant on a large number of keypoints. Although significant decreases in performance occur when using 8 keypoints, we see small differences for larger amounts, with the optimal number of keypoints being 64 for both datasets.} 
  \label{fig:keypoints}
\end{figure}

\subsection{Ablation studies}

\noindent\textbf{Ablation study on number of \textbf{OSKFs}} On Fig. \ref{fig:keypoints} we show that \Ours~is not highly reliant in a high number of reference keypoints. While we find that 8 points does not offer sufficient accuracy for our standards, the difference between 16 and 128 keypoints is not significant. Regardless, we use $K=64$ keypoints on all our experiments as it provides the best results for both LM-O and YCB-V.

\noindent\textbf{Cascaded Pose Refinement.} We proposed a method that iteratively refines the initial pose of an object in a cascaded fashion. On Tab. \ref{tab:refinement_ablation} we show the impact multiple iterations have on the accuracy. We find a decrease of marginal improvements when applying multiple refinements. The reduction in error on the first refinement iteration is larger than the consequent iterations combined.

\noindent\textbf{Effect of coarse pose accuracy. }  In Table \ref{tab:initial_ablation} we show the impact of the initial coarse pose on the OSKF-TP module. 
We show OSKF-TP can serve as an independent refiner module, such as DeepIM~\cite{deepim}  or Repose~\cite{repose}, as it can receive pose estimations from other approaches and refine them to a state of the art level. Using 3 OSKF-TP modules, we improve GDR~\cite{gdr} and SO-Pose~\cite{sopose} results by $22\%$ and $10\%$ respectively. This experiment is done for the sake of completion: \Ours~ must generate the feature pyramid $\mathcal{F}$ to compute OSKFs, which is the most expensive operation in the pipeline, making the use of these initial poses redundant and unnecessary as we achieve better results with our coarse initialization.

\begin{table}[]
\resizebox{\columnwidth}{!}{%
\begin{tabular}{c|cc|ccc}
\Xhline{5\arrayrulewidth}
Dataset & \multicolumn{2}{c|}{LM-O}       & \multicolumn{3}{c}{YCB-V}                                    \\ \hline
Metric  & \multicolumn{1}{c|}{ADD-(S)}   & AR & \multicolumn{1}{c|}{ADD-(S)}  & \multicolumn{1}{c|}{AUC}  & AR   \\ \hline \hline
0       & \multicolumn{1}{c|}{40.5}  &  59.8  & \multicolumn{1}{c|}{50.6} & \multicolumn{1}{c|}{81.0} & 59.7 \\ \hline
1       & \multicolumn{1}{c|}{61.4} &  69.0  & \multicolumn{1}{c|}{63.2} & \multicolumn{1}{c|}{84.9} & 70.2 \\ \hline
2       & \multicolumn{1}{c|}{64.9} &  71.1  & \multicolumn{1}{c|}{70.8} & \multicolumn{1}{c|}{87.1} & 75.2 \\ \hline
3       & \multicolumn{1}{c|}{66.3}  &  71.5  & \multicolumn{1}{c|}{72.1} & \multicolumn{1}{c|}{87.5} & 76.3 \\
\Xhline{5\arrayrulewidth}
\end{tabular}
}
\caption{\textbf{Effects of the Cascaded Pose Refinement}. We present the improvement with each refinement step. We can observe the diminishing returns where the first refinement improvement is larger than all other iterations combined.}
\label{tab:refinement_ablation}
\vspace{-5pt}
\end{table}

\begin{table}[t]
\resizebox{\columnwidth}{!}{%
\begin{tabular}{@{}c|ccc|c|c|c@{}} 
\Xhline{5\arrayrulewidth}
\multirow{2}{*}{Initial Pose} & \multicolumn{3}{c|}{ADD(-S)}                                   & \multirow{2}{*}{2°2cm} & \multirow{2}{*}{5°5cm} & \multirow{2}{*}{Total} \\ \cline{2-4}
                              & \multicolumn{1}{c|}{0.02d} & \multicolumn{1}{c|}{0.05d} & 0.1d &                        &                        &                        \\ \hline
GDR~\cite{gdr}                           & \multicolumn{1}{c|}{3.42}  & \multicolumn{1}{c|}{26.5}  & 56.1 & 3.14                   & 35.1                   & 24.9                   \\ \hline
SO-Pose~\cite{sopose}                       & \multicolumn{1}{c|}{\textbf{4.55}}  & \multicolumn{1}{c|}{\textbf{31.4}}  & \textbf{62.3 }& \textbf{3.39}                   & \textbf{39.0}                   & \textbf{28.1}                   \\ \hline
\textbf{\Ours-Coarse}                        & \multicolumn{1}{c|}{1.53}  & \multicolumn{1}{c|}{16.3}  & 40.5 & 1.99                   & 29.45                  & 18.0                   \\ \hline \hline
GDR$^*$~\cite{gdr}                          & \multicolumn{1}{c|}{6.08}  & \multicolumn{1}{c|}{34.4}  & 62.5 & 5.67                   & 43.5                   & 30.43                  \\ \hline
SO-POSE$^*$~\cite{sopose}                     & \multicolumn{1}{c|}{6.22}  & \multicolumn{1}{c|}{34.8}  & 63.4 & 5.38                   & 44.4                   & 30.84                  \\ \hline
\textbf{\Ours}                        & \multicolumn{1}{c|}{\textbf{6.23}}  & \multicolumn{1}{c|}{\textbf{36.2}}  &\textbf{ 66.3 }& \textbf{5.48}                   & \textbf{45.4}                   & \textbf{31.9}    \\

\Xhline{5\arrayrulewidth}
\end{tabular}
}
\vspace{2pt}
\caption{\textbf{Initial Pose ablation study.} We show the experimental results when using our refinement on top of prior art. Results for initial poses are on the top 3 rows while bottom 3 rows are the refined poses. * indicates the use of \Ours~refinement on the respective method's estimated pose.}
\label{tab:initial_ablation}
\vspace{-15pt}
\end{table}



\subsection{Qualitative Results}

\noindent \textbf{Visualization of Deformable Attention.}
In Fig \ref{fig:attention} we visualize the attention sampling points generated by the deformable attention operation. We only show sampling positions (white circles) for high attention weights $A>0.25$ (see Eq.~\ref{eq:attention_deform}). When the keypoints is visible, the deformations occur near the keypoint projection location, whereas for occluded keypoints attention is scattered around the image. The transformer can guide its sampling to regions with clues to recover subtle pose differences.

\begin{figure}[t]
  \vspace{-5pt}
    \centering 
    \begin{subfigure}{0.18\columnwidth}
      \includegraphics[width=\linewidth]{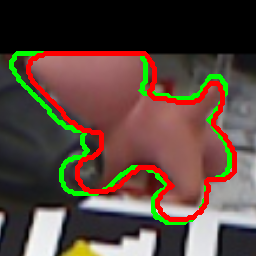}
      \label{}
    \end{subfigure}\hfil  
    \begin{subfigure}{0.18\columnwidth}
      \includegraphics[width=\linewidth]{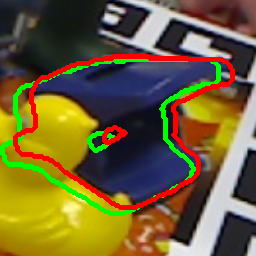}
      \label{}
    \end{subfigure}\hfil  
    \begin{subfigure}{0.18\columnwidth}
      \includegraphics[width=\linewidth]{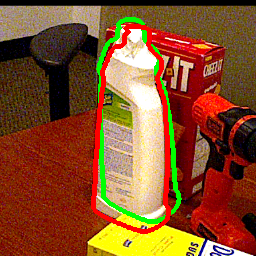}
      \label{}
    \end{subfigure}\hfil 
    \begin{subfigure}{0.18\columnwidth}
      \includegraphics[width=\linewidth]{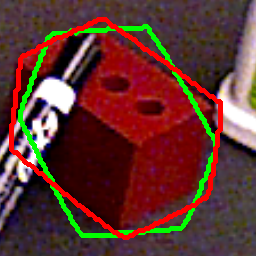}
      \label{}
    \end{subfigure}\hfil 
    \begin{subfigure}{0.18\columnwidth}
      \includegraphics[width=\linewidth]{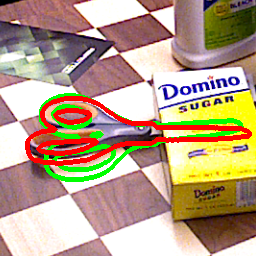}
      \label{}
    \end{subfigure}  \hfil 
    
    \vspace{-5pt}
    \begin{subfigure}{0.18\columnwidth}
      \includegraphics[width=\linewidth]{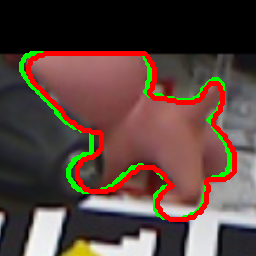}
      \label{}
    \end{subfigure}\hfil 
    \begin{subfigure}{0.18\columnwidth}
      \includegraphics[width=\linewidth]{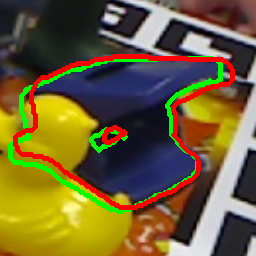}
      \label{}
    \end{subfigure}\hfil  
    \begin{subfigure}{0.18\columnwidth}
      \includegraphics[width=\linewidth]{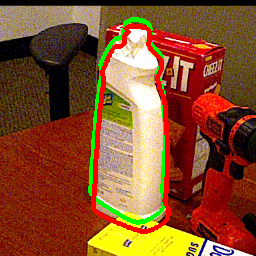}
      \label{}
    \end{subfigure}\hfil 
    \begin{subfigure}{0.18\columnwidth}
      \includegraphics[width=\linewidth]{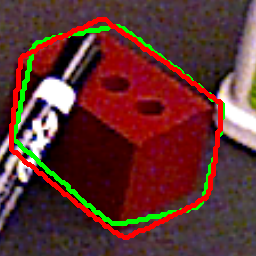}
      \label{}
    \end{subfigure}\hfil 
    \begin{subfigure}{0.18\columnwidth}
      \includegraphics[width=\linewidth]{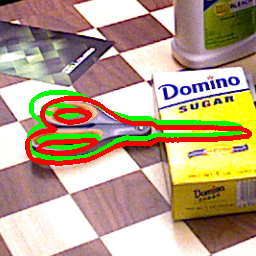}
      \label{}
    \end{subfigure}\hfil 
    
    \vspace{-5pt}
    \begin{subfigure}{0.18\columnwidth}
      \includegraphics[width=\linewidth]{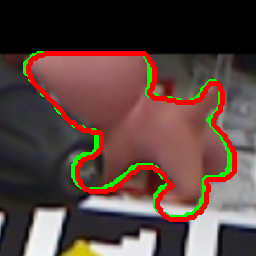}
      \label{}
    \end{subfigure}\hfil 
    \begin{subfigure}{0.18\columnwidth}
      \includegraphics[width=\linewidth]{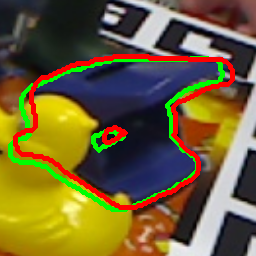}
      \label{}
    \end{subfigure}\hfil  
    \begin{subfigure}{0.18\columnwidth}
      \includegraphics[width=\linewidth]{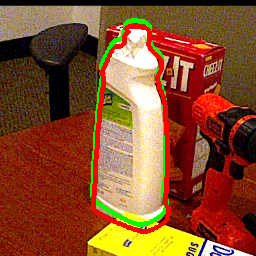}
      \label{}
    \end{subfigure}\hfil 
    \begin{subfigure}{0.18\columnwidth}
      \includegraphics[width=\linewidth]{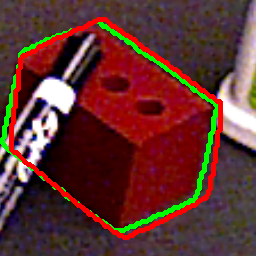}
      \label{}
    \end{subfigure}\hfil 
    \begin{subfigure}{0.18\columnwidth}
      \includegraphics[width=\linewidth]{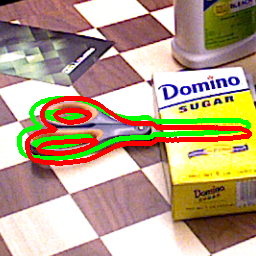}
      \label{}
    \end{subfigure}\hfil 
    \vspace{-5pt}
    \noindent\rule{8cm}{0.4pt}
    \\
    \medskip
    \begin{subfigure}{0.18\columnwidth}
      \includegraphics[width=\linewidth]{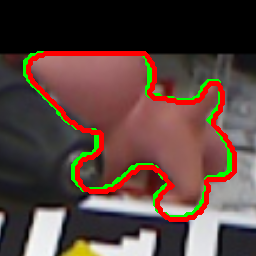}
      \label{}
    \end{subfigure}\hfil 
    \begin{subfigure}{0.18\columnwidth}
      \includegraphics[width=\linewidth]{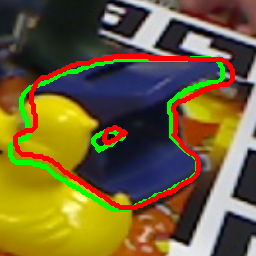}
      \label{}
    \end{subfigure}\hfil  
    \begin{subfigure}{0.18\columnwidth}
      \includegraphics[width=\linewidth]{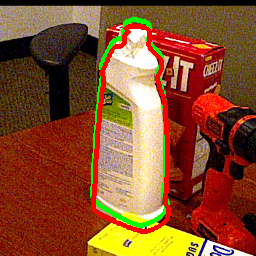}
      \label{}
    \end{subfigure}\hfil 
    \begin{subfigure}{0.18\columnwidth}
      \includegraphics[width=\linewidth]{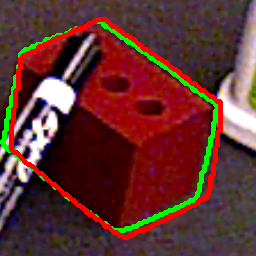}
      \label{}
    \end{subfigure}\hfil 
    \begin{subfigure}{0.18\columnwidth}
      \includegraphics[width=\linewidth]{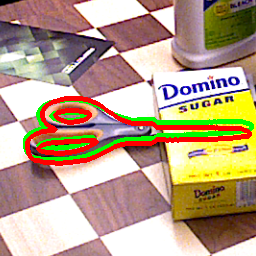}
      \label{}
    \end{subfigure}
\vspace{-15pt}
\caption{\textbf{Qualitative results on refinement steps.} We show the progress in the pose accuracy over the iterative refinements, with the first row being the initial coarse pose and last row the final results. Although the initial pose is very poor for evaluation standards it proves to suffice as a reference for refinement.}
\label{fig:images}
\vspace{-10pt}
\end{figure}

\noindent \textbf{Refinement Qualitative Examples.} 
On Fig.\ref{fig:images} we present qualitative results showing the impact of each refinement step.
We can see that for most objects the original pose is significantly inaccurate. Nonetheless, \Ours~can recover an accurate pose after 3 iterations of the refinement module. 
The last column shows the results on the \textit{scissors} object, the hardest object in YCB-V, \Ours~was not able to recover the correct pose due to a poor initial pose.

%% file: resources/figure.pgf
\begingroup%
\makeatletter%
\begin{pgfpicture}%
\pgfpathrectangle{\pgfpointorigin}{\pgfqpoint{5.535208in}{2.250369in}}%
\pgfusepath{use as bounding box, clip}%
\begin{pgfscope}%
\pgfsetbuttcap%
\pgfsetmiterjoin%
\definecolor{currentfill}{rgb}{1.000000,1.000000,1.000000}%
\pgfsetfillcolor{currentfill}%
\pgfsetlinewidth{0.000000pt}%
\definecolor{currentstroke}{rgb}{1.000000,1.000000,1.000000}%
\pgfsetstrokecolor{currentstroke}%
\pgfsetdash{}{0pt}%
\pgfpathmoveto{\pgfqpoint{0.000000in}{0.000000in}}%
\pgfpathlineto{\pgfqpoint{5.535208in}{0.000000in}}%
\pgfpathlineto{\pgfqpoint{5.535208in}{2.250369in}}%
\pgfpathlineto{\pgfqpoint{0.000000in}{2.250369in}}%
\pgfpathclose%
\pgfusepath{fill}%
\end{pgfscope}%
\begin{pgfscope}%
\pgfsetbuttcap%
\pgfsetmiterjoin%
\definecolor{currentfill}{rgb}{1.000000,1.000000,1.000000}%
\pgfsetfillcolor{currentfill}%
\pgfsetlinewidth{0.000000pt}%
\definecolor{currentstroke}{rgb}{0.000000,0.000000,0.000000}%
\pgfsetstrokecolor{currentstroke}%
\pgfsetstrokeopacity{0.000000}%
\pgfsetdash{}{0pt}%
\pgfpathmoveto{\pgfqpoint{0.626150in}{0.575369in}}%
\pgfpathlineto{\pgfqpoint{5.276150in}{0.575369in}}%
\pgfpathlineto{\pgfqpoint{5.276150in}{2.115369in}}%
\pgfpathlineto{\pgfqpoint{0.626150in}{2.115369in}}%
\pgfpathclose%
\pgfusepath{fill}%
\end{pgfscope}%
\begin{pgfscope}%
\pgfsetbuttcap%
\pgfsetroundjoin%
\definecolor{currentfill}{rgb}{0.000000,0.000000,0.000000}%
\pgfsetfillcolor{currentfill}%
\pgfsetlinewidth{0.803000pt}%
\definecolor{currentstroke}{rgb}{0.000000,0.000000,0.000000}%
\pgfsetstrokecolor{currentstroke}%
\pgfsetdash{}{0pt}%
\pgfsys@defobject{currentmarker}{\pgfqpoint{0.000000in}{-0.048611in}}{\pgfqpoint{0.000000in}{0.000000in}}{%
\pgfpathmoveto{\pgfqpoint{0.000000in}{0.000000in}}%
\pgfpathlineto{\pgfqpoint{0.000000in}{-0.048611in}}%
\pgfusepath{stroke,fill}%
}%
\begin{pgfscope}%
\pgfsys@transformshift{0.626150in}{0.575369in}%
\pgfsys@useobject{currentmarker}{}%
\end{pgfscope}%
\end{pgfscope}%
\begin{pgfscope}%
\pgftext[x=0.626150in,y=0.478146in,,top]{\sffamily\fontsize{12.000000}{14.400000}\selectfont 0}%
\end{pgfscope}%
\begin{pgfscope}%
\pgfsetbuttcap%
\pgfsetroundjoin%
\definecolor{currentfill}{rgb}{0.000000,0.000000,0.000000}%
\pgfsetfillcolor{currentfill}%
\pgfsetlinewidth{0.803000pt}%
\definecolor{currentstroke}{rgb}{0.000000,0.000000,0.000000}%
\pgfsetstrokecolor{currentstroke}%
\pgfsetdash{}{0pt}%
\pgfsys@defobject{currentmarker}{\pgfqpoint{0.000000in}{-0.048611in}}{\pgfqpoint{0.000000in}{0.000000in}}{%
\pgfpathmoveto{\pgfqpoint{0.000000in}{0.000000in}}%
\pgfpathlineto{\pgfqpoint{0.000000in}{-0.048611in}}%
\pgfusepath{stroke,fill}%
}%
\begin{pgfscope}%
\pgfsys@transformshift{1.556150in}{0.575369in}%
\pgfsys@useobject{currentmarker}{}%
\end{pgfscope}%
\end{pgfscope}%
\begin{pgfscope}%
\pgftext[x=1.556150in,y=0.478146in,,top]{\sffamily\fontsize{12.000000}{14.400000}\selectfont 100}%
\end{pgfscope}%
\begin{pgfscope}%
\pgfsetbuttcap%
\pgfsetroundjoin%
\definecolor{currentfill}{rgb}{0.000000,0.000000,0.000000}%
\pgfsetfillcolor{currentfill}%
\pgfsetlinewidth{0.803000pt}%
\definecolor{currentstroke}{rgb}{0.000000,0.000000,0.000000}%
\pgfsetstrokecolor{currentstroke}%
\pgfsetdash{}{0pt}%
\pgfsys@defobject{currentmarker}{\pgfqpoint{0.000000in}{-0.048611in}}{\pgfqpoint{0.000000in}{0.000000in}}{%
\pgfpathmoveto{\pgfqpoint{0.000000in}{0.000000in}}%
\pgfpathlineto{\pgfqpoint{0.000000in}{-0.048611in}}%
\pgfusepath{stroke,fill}%
}%
\begin{pgfscope}%
\pgfsys@transformshift{2.486150in}{0.575369in}%
\pgfsys@useobject{currentmarker}{}%
\end{pgfscope}%
\end{pgfscope}%
\begin{pgfscope}%
\pgftext[x=2.486150in,y=0.478146in,,top]{\sffamily\fontsize{12.000000}{14.400000}\selectfont 200}%
\end{pgfscope}%
\begin{pgfscope}%
\pgfsetbuttcap%
\pgfsetroundjoin%
\definecolor{currentfill}{rgb}{0.000000,0.000000,0.000000}%
\pgfsetfillcolor{currentfill}%
\pgfsetlinewidth{0.803000pt}%
\definecolor{currentstroke}{rgb}{0.000000,0.000000,0.000000}%
\pgfsetstrokecolor{currentstroke}%
\pgfsetdash{}{0pt}%
\pgfsys@defobject{currentmarker}{\pgfqpoint{0.000000in}{-0.048611in}}{\pgfqpoint{0.000000in}{0.000000in}}{%
\pgfpathmoveto{\pgfqpoint{0.000000in}{0.000000in}}%
\pgfpathlineto{\pgfqpoint{0.000000in}{-0.048611in}}%
\pgfusepath{stroke,fill}%
}%
\begin{pgfscope}%
\pgfsys@transformshift{3.416150in}{0.575369in}%
\pgfsys@useobject{currentmarker}{}%
\end{pgfscope}%
\end{pgfscope}%
\begin{pgfscope}%
\pgftext[x=3.416150in,y=0.478146in,,top]{\sffamily\fontsize{12.000000}{14.400000}\selectfont 300}%
\end{pgfscope}%
\begin{pgfscope}%
\pgfsetbuttcap%
\pgfsetroundjoin%
\definecolor{currentfill}{rgb}{0.000000,0.000000,0.000000}%
\pgfsetfillcolor{currentfill}%
\pgfsetlinewidth{0.803000pt}%
\definecolor{currentstroke}{rgb}{0.000000,0.000000,0.000000}%
\pgfsetstrokecolor{currentstroke}%
\pgfsetdash{}{0pt}%
\pgfsys@defobject{currentmarker}{\pgfqpoint{0.000000in}{-0.048611in}}{\pgfqpoint{0.000000in}{0.000000in}}{%
\pgfpathmoveto{\pgfqpoint{0.000000in}{0.000000in}}%
\pgfpathlineto{\pgfqpoint{0.000000in}{-0.048611in}}%
\pgfusepath{stroke,fill}%
}%
\begin{pgfscope}%
\pgfsys@transformshift{4.346150in}{0.575369in}%
\pgfsys@useobject{currentmarker}{}%
\end{pgfscope}%
\end{pgfscope}%
\begin{pgfscope}%
\pgftext[x=4.346150in,y=0.478146in,,top]{\sffamily\fontsize{12.000000}{14.400000}\selectfont 400}%
\end{pgfscope}%
\begin{pgfscope}%
\pgfsetbuttcap%
\pgfsetroundjoin%
\definecolor{currentfill}{rgb}{0.000000,0.000000,0.000000}%
\pgfsetfillcolor{currentfill}%
\pgfsetlinewidth{0.803000pt}%
\definecolor{currentstroke}{rgb}{0.000000,0.000000,0.000000}%
\pgfsetstrokecolor{currentstroke}%
\pgfsetdash{}{0pt}%
\pgfsys@defobject{currentmarker}{\pgfqpoint{0.000000in}{-0.048611in}}{\pgfqpoint{0.000000in}{0.000000in}}{%
\pgfpathmoveto{\pgfqpoint{0.000000in}{0.000000in}}%
\pgfpathlineto{\pgfqpoint{0.000000in}{-0.048611in}}%
\pgfusepath{stroke,fill}%
}%
\begin{pgfscope}%
\pgfsys@transformshift{5.276150in}{0.575369in}%
\pgfsys@useobject{currentmarker}{}%
\end{pgfscope}%
\end{pgfscope}%
\begin{pgfscope}%
\pgftext[x=5.276150in,y=0.478146in,,top]{\sffamily\fontsize{12.000000}{14.400000}\selectfont 500}%
\end{pgfscope}%
\begin{pgfscope}%
\pgftext[x=2.951150in,y=0.261295in,,top]{\sffamily\fontsize{12.000000}{14.400000}\selectfont Time(ms)/image}%
\end{pgfscope}%
\begin{pgfscope}%
\pgfsetbuttcap%
\pgfsetroundjoin%
\definecolor{currentfill}{rgb}{0.000000,0.000000,0.000000}%
\pgfsetfillcolor{currentfill}%
\pgfsetlinewidth{0.803000pt}%
\definecolor{currentstroke}{rgb}{0.000000,0.000000,0.000000}%
\pgfsetstrokecolor{currentstroke}%
\pgfsetdash{}{0pt}%
\pgfsys@defobject{currentmarker}{\pgfqpoint{-0.048611in}{0.000000in}}{\pgfqpoint{0.000000in}{0.000000in}}{%
\pgfpathmoveto{\pgfqpoint{0.000000in}{0.000000in}}%
\pgfpathlineto{\pgfqpoint{-0.048611in}{0.000000in}}%
\pgfusepath{stroke,fill}%
}%
\begin{pgfscope}%
\pgfsys@transformshift{0.626150in}{0.575369in}%
\pgfsys@useobject{currentmarker}{}%
\end{pgfscope}%
\end{pgfscope}%
\begin{pgfscope}%
\pgftext[x=0.316851in,y=0.512055in,left,base]{\sffamily\fontsize{12.000000}{14.400000}\selectfont 50}%
\end{pgfscope}%
\begin{pgfscope}%
\pgfsetbuttcap%
\pgfsetroundjoin%
\definecolor{currentfill}{rgb}{0.000000,0.000000,0.000000}%
\pgfsetfillcolor{currentfill}%
\pgfsetlinewidth{0.803000pt}%
\definecolor{currentstroke}{rgb}{0.000000,0.000000,0.000000}%
\pgfsetstrokecolor{currentstroke}%
\pgfsetdash{}{0pt}%
\pgfsys@defobject{currentmarker}{\pgfqpoint{-0.048611in}{0.000000in}}{\pgfqpoint{0.000000in}{0.000000in}}{%
\pgfpathmoveto{\pgfqpoint{0.000000in}{0.000000in}}%
\pgfpathlineto{\pgfqpoint{-0.048611in}{0.000000in}}%
\pgfusepath{stroke,fill}%
}%
\begin{pgfscope}%
\pgfsys@transformshift{0.626150in}{1.015369in}%
\pgfsys@useobject{currentmarker}{}%
\end{pgfscope}%
\end{pgfscope}%
\begin{pgfscope}%
\pgftext[x=0.316851in,y=0.952055in,left,base]{\sffamily\fontsize{12.000000}{14.400000}\selectfont 60}%
\end{pgfscope}%
\begin{pgfscope}%
\pgfsetbuttcap%
\pgfsetroundjoin%
\definecolor{currentfill}{rgb}{0.000000,0.000000,0.000000}%
\pgfsetfillcolor{currentfill}%
\pgfsetlinewidth{0.803000pt}%
\definecolor{currentstroke}{rgb}{0.000000,0.000000,0.000000}%
\pgfsetstrokecolor{currentstroke}%
\pgfsetdash{}{0pt}%
\pgfsys@defobject{currentmarker}{\pgfqpoint{-0.048611in}{0.000000in}}{\pgfqpoint{0.000000in}{0.000000in}}{%
\pgfpathmoveto{\pgfqpoint{0.000000in}{0.000000in}}%
\pgfpathlineto{\pgfqpoint{-0.048611in}{0.000000in}}%
\pgfusepath{stroke,fill}%
}%
\begin{pgfscope}%
\pgfsys@transformshift{0.626150in}{1.455369in}%
\pgfsys@useobject{currentmarker}{}%
\end{pgfscope}%
\end{pgfscope}%
\begin{pgfscope}%
\pgftext[x=0.316851in,y=1.392055in,left,base]{\sffamily\fontsize{12.000000}{14.400000}\selectfont 70}%
\end{pgfscope}%
\begin{pgfscope}%
\pgfsetbuttcap%
\pgfsetroundjoin%
\definecolor{currentfill}{rgb}{0.000000,0.000000,0.000000}%
\pgfsetfillcolor{currentfill}%
\pgfsetlinewidth{0.803000pt}%
\definecolor{currentstroke}{rgb}{0.000000,0.000000,0.000000}%
\pgfsetstrokecolor{currentstroke}%
\pgfsetdash{}{0pt}%
\pgfsys@defobject{currentmarker}{\pgfqpoint{-0.048611in}{0.000000in}}{\pgfqpoint{0.000000in}{0.000000in}}{%
\pgfpathmoveto{\pgfqpoint{0.000000in}{0.000000in}}%
\pgfpathlineto{\pgfqpoint{-0.048611in}{0.000000in}}%
\pgfusepath{stroke,fill}%
}%
\begin{pgfscope}%
\pgfsys@transformshift{0.626150in}{1.895369in}%
\pgfsys@useobject{currentmarker}{}%
\end{pgfscope}%
\end{pgfscope}%
\begin{pgfscope}%
\pgftext[x=0.316851in,y=1.832055in,left,base]{\sffamily\fontsize{12.000000}{14.400000}\selectfont 80}%
\end{pgfscope}%
\begin{pgfscope}%
\pgftext[x=0.261295in,y=1.345369in,,bottom,rotate=90.000000]{\sffamily\fontsize{12.000000}{14.400000}\selectfont Average Recall}%
\end{pgfscope}%
\begin{pgfscope}%
\pgfpathrectangle{\pgfqpoint{0.626150in}{0.575369in}}{\pgfqpoint{4.650000in}{1.540000in}}%
\pgfusepath{clip}%
\pgfsetbuttcap%
\pgfsetroundjoin%
\pgfsetlinewidth{1.505625pt}%
\definecolor{currentstroke}{rgb}{1.000000,0.000000,0.000000}%
\pgfsetstrokecolor{currentstroke}%
\pgfsetdash{{5.550000pt}{2.400000pt}}{0.000000pt}%
\pgfpathmoveto{\pgfqpoint{0.830750in}{1.002169in}}%
\pgfpathlineto{\pgfqpoint{0.844700in}{1.464169in}}%
\pgfpathlineto{\pgfqpoint{0.858650in}{1.684169in}}%
\pgfpathlineto{\pgfqpoint{0.872600in}{1.732569in}}%
\pgfusepath{stroke}%
\end{pgfscope}%
\begin{pgfscope}%
\pgfpathrectangle{\pgfqpoint{0.626150in}{0.575369in}}{\pgfqpoint{4.650000in}{1.540000in}}%
\pgfusepath{clip}%
\pgfsetbuttcap%
\pgfsetmiterjoin%
\definecolor{currentfill}{rgb}{1.000000,0.000000,0.000000}%
\pgfsetfillcolor{currentfill}%
\pgfsetlinewidth{1.003750pt}%
\definecolor{currentstroke}{rgb}{1.000000,0.000000,0.000000}%
\pgfsetstrokecolor{currentstroke}%
\pgfsetdash{}{0pt}%
\pgfsys@defobject{currentmarker}{\pgfqpoint{-0.041667in}{-0.041667in}}{\pgfqpoint{0.041667in}{0.041667in}}{%
\pgfpathmoveto{\pgfqpoint{0.000000in}{0.041667in}}%
\pgfpathlineto{\pgfqpoint{-0.041667in}{-0.041667in}}%
\pgfpathlineto{\pgfqpoint{0.041667in}{-0.041667in}}%
\pgfpathclose%
\pgfusepath{stroke,fill}%
}%
\begin{pgfscope}%
\pgfsys@transformshift{0.830750in}{1.002169in}%
\pgfsys@useobject{currentmarker}{}%
\end{pgfscope}%
\begin{pgfscope}%
\pgfsys@transformshift{0.844700in}{1.464169in}%
\pgfsys@useobject{currentmarker}{}%
\end{pgfscope}%
\begin{pgfscope}%
\pgfsys@transformshift{0.858650in}{1.684169in}%
\pgfsys@useobject{currentmarker}{}%
\end{pgfscope}%
\begin{pgfscope}%
\pgfsys@transformshift{0.872600in}{1.732569in}%
\pgfsys@useobject{currentmarker}{}%
\end{pgfscope}%
\end{pgfscope}%
\begin{pgfscope}%
\pgfpathrectangle{\pgfqpoint{0.626150in}{0.575369in}}{\pgfqpoint{4.650000in}{1.540000in}}%
\pgfusepath{clip}%
\pgfsetrectcap%
\pgfsetroundjoin%
\pgfsetlinewidth{1.505625pt}%
\definecolor{currentstroke}{rgb}{0.000000,0.000000,0.000000}%
\pgfsetstrokecolor{currentstroke}%
\pgfsetdash{}{0pt}%
\pgfpathmoveto{\pgfqpoint{1.249250in}{1.516969in}}%
\pgfusepath{stroke}%
\end{pgfscope}%
\begin{pgfscope}%
\pgfpathrectangle{\pgfqpoint{0.626150in}{0.575369in}}{\pgfqpoint{4.650000in}{1.540000in}}%
\pgfusepath{clip}%
\pgfsetbuttcap%
\pgfsetroundjoin%
\definecolor{currentfill}{rgb}{0.000000,0.000000,0.000000}%
\pgfsetfillcolor{currentfill}%
\pgfsetlinewidth{1.003750pt}%
\definecolor{currentstroke}{rgb}{0.000000,0.000000,0.000000}%
\pgfsetstrokecolor{currentstroke}%
\pgfsetdash{}{0pt}%
\pgfsys@defobject{currentmarker}{\pgfqpoint{-0.020833in}{-0.020833in}}{\pgfqpoint{0.020833in}{0.020833in}}{%
\pgfpathmoveto{\pgfqpoint{0.000000in}{-0.020833in}}%
\pgfpathcurveto{\pgfqpoint{0.005525in}{-0.020833in}}{\pgfqpoint{0.010825in}{-0.018638in}}{\pgfqpoint{0.014731in}{-0.014731in}}%
\pgfpathcurveto{\pgfqpoint{0.018638in}{-0.010825in}}{\pgfqpoint{0.020833in}{-0.005525in}}{\pgfqpoint{0.020833in}{0.000000in}}%
\pgfpathcurveto{\pgfqpoint{0.020833in}{0.005525in}}{\pgfqpoint{0.018638in}{0.010825in}}{\pgfqpoint{0.014731in}{0.014731in}}%
\pgfpathcurveto{\pgfqpoint{0.010825in}{0.018638in}}{\pgfqpoint{0.005525in}{0.020833in}}{\pgfqpoint{0.000000in}{0.020833in}}%
\pgfpathcurveto{\pgfqpoint{-0.005525in}{0.020833in}}{\pgfqpoint{-0.010825in}{0.018638in}}{\pgfqpoint{-0.014731in}{0.014731in}}%
\pgfpathcurveto{\pgfqpoint{-0.018638in}{0.010825in}}{\pgfqpoint{-0.020833in}{0.005525in}}{\pgfqpoint{-0.020833in}{0.000000in}}%
\pgfpathcurveto{\pgfqpoint{-0.020833in}{-0.005525in}}{\pgfqpoint{-0.018638in}{-0.010825in}}{\pgfqpoint{-0.014731in}{-0.014731in}}%
\pgfpathcurveto{\pgfqpoint{-0.010825in}{-0.018638in}}{\pgfqpoint{-0.005525in}{-0.020833in}}{\pgfqpoint{0.000000in}{-0.020833in}}%
\pgfpathclose%
\pgfusepath{stroke,fill}%
}%
\begin{pgfscope}%
\pgfsys@transformshift{1.249250in}{1.516969in}%
\pgfsys@useobject{currentmarker}{}%
\end{pgfscope}%
\end{pgfscope}%
\begin{pgfscope}%
\pgfpathrectangle{\pgfqpoint{0.626150in}{0.575369in}}{\pgfqpoint{4.650000in}{1.540000in}}%
\pgfusepath{clip}%
\pgfsetrectcap%
\pgfsetroundjoin%
\pgfsetlinewidth{1.505625pt}%
\definecolor{currentstroke}{rgb}{0.000000,0.000000,0.000000}%
\pgfsetstrokecolor{currentstroke}%
\pgfsetdash{}{0pt}%
\pgfpathmoveto{\pgfqpoint{1.081850in}{1.283769in}}%
\pgfusepath{stroke}%
\end{pgfscope}%
\begin{pgfscope}%
\pgfpathrectangle{\pgfqpoint{0.626150in}{0.575369in}}{\pgfqpoint{4.650000in}{1.540000in}}%
\pgfusepath{clip}%
\pgfsetbuttcap%
\pgfsetroundjoin%
\definecolor{currentfill}{rgb}{0.000000,0.000000,0.000000}%
\pgfsetfillcolor{currentfill}%
\pgfsetlinewidth{1.003750pt}%
\definecolor{currentstroke}{rgb}{0.000000,0.000000,0.000000}%
\pgfsetstrokecolor{currentstroke}%
\pgfsetdash{}{0pt}%
\pgfsys@defobject{currentmarker}{\pgfqpoint{-0.020833in}{-0.020833in}}{\pgfqpoint{0.020833in}{0.020833in}}{%
\pgfpathmoveto{\pgfqpoint{0.000000in}{-0.020833in}}%
\pgfpathcurveto{\pgfqpoint{0.005525in}{-0.020833in}}{\pgfqpoint{0.010825in}{-0.018638in}}{\pgfqpoint{0.014731in}{-0.014731in}}%
\pgfpathcurveto{\pgfqpoint{0.018638in}{-0.010825in}}{\pgfqpoint{0.020833in}{-0.005525in}}{\pgfqpoint{0.020833in}{0.000000in}}%
\pgfpathcurveto{\pgfqpoint{0.020833in}{0.005525in}}{\pgfqpoint{0.018638in}{0.010825in}}{\pgfqpoint{0.014731in}{0.014731in}}%
\pgfpathcurveto{\pgfqpoint{0.010825in}{0.018638in}}{\pgfqpoint{0.005525in}{0.020833in}}{\pgfqpoint{0.000000in}{0.020833in}}%
\pgfpathcurveto{\pgfqpoint{-0.005525in}{0.020833in}}{\pgfqpoint{-0.010825in}{0.018638in}}{\pgfqpoint{-0.014731in}{0.014731in}}%
\pgfpathcurveto{\pgfqpoint{-0.018638in}{0.010825in}}{\pgfqpoint{-0.020833in}{0.005525in}}{\pgfqpoint{-0.020833in}{0.000000in}}%
\pgfpathcurveto{\pgfqpoint{-0.020833in}{-0.005525in}}{\pgfqpoint{-0.018638in}{-0.010825in}}{\pgfqpoint{-0.014731in}{-0.014731in}}%
\pgfpathcurveto{\pgfqpoint{-0.010825in}{-0.018638in}}{\pgfqpoint{-0.005525in}{-0.020833in}}{\pgfqpoint{0.000000in}{-0.020833in}}%
\pgfpathclose%
\pgfusepath{stroke,fill}%
}%
\begin{pgfscope}%
\pgfsys@transformshift{1.081850in}{1.283769in}%
\pgfsys@useobject{currentmarker}{}%
\end{pgfscope}%
\end{pgfscope}%
\begin{pgfscope}%
\pgfpathrectangle{\pgfqpoint{0.626150in}{0.575369in}}{\pgfqpoint{4.650000in}{1.540000in}}%
\pgfusepath{clip}%
\pgfsetrectcap%
\pgfsetroundjoin%
\pgfsetlinewidth{1.505625pt}%
\definecolor{currentstroke}{rgb}{0.000000,0.000000,0.000000}%
\pgfsetstrokecolor{currentstroke}%
\pgfsetdash{}{0pt}%
\pgfpathmoveto{\pgfqpoint{1.332950in}{0.716169in}}%
\pgfusepath{stroke}%
\end{pgfscope}%
\begin{pgfscope}%
\pgfpathrectangle{\pgfqpoint{0.626150in}{0.575369in}}{\pgfqpoint{4.650000in}{1.540000in}}%
\pgfusepath{clip}%
\pgfsetbuttcap%
\pgfsetroundjoin%
\definecolor{currentfill}{rgb}{0.000000,0.000000,0.000000}%
\pgfsetfillcolor{currentfill}%
\pgfsetlinewidth{1.003750pt}%
\definecolor{currentstroke}{rgb}{0.000000,0.000000,0.000000}%
\pgfsetstrokecolor{currentstroke}%
\pgfsetdash{}{0pt}%
\pgfsys@defobject{currentmarker}{\pgfqpoint{-0.020833in}{-0.020833in}}{\pgfqpoint{0.020833in}{0.020833in}}{%
\pgfpathmoveto{\pgfqpoint{0.000000in}{-0.020833in}}%
\pgfpathcurveto{\pgfqpoint{0.005525in}{-0.020833in}}{\pgfqpoint{0.010825in}{-0.018638in}}{\pgfqpoint{0.014731in}{-0.014731in}}%
\pgfpathcurveto{\pgfqpoint{0.018638in}{-0.010825in}}{\pgfqpoint{0.020833in}{-0.005525in}}{\pgfqpoint{0.020833in}{0.000000in}}%
\pgfpathcurveto{\pgfqpoint{0.020833in}{0.005525in}}{\pgfqpoint{0.018638in}{0.010825in}}{\pgfqpoint{0.014731in}{0.014731in}}%
\pgfpathcurveto{\pgfqpoint{0.010825in}{0.018638in}}{\pgfqpoint{0.005525in}{0.020833in}}{\pgfqpoint{0.000000in}{0.020833in}}%
\pgfpathcurveto{\pgfqpoint{-0.005525in}{0.020833in}}{\pgfqpoint{-0.010825in}{0.018638in}}{\pgfqpoint{-0.014731in}{0.014731in}}%
\pgfpathcurveto{\pgfqpoint{-0.018638in}{0.010825in}}{\pgfqpoint{-0.020833in}{0.005525in}}{\pgfqpoint{-0.020833in}{0.000000in}}%
\pgfpathcurveto{\pgfqpoint{-0.020833in}{-0.005525in}}{\pgfqpoint{-0.018638in}{-0.010825in}}{\pgfqpoint{-0.014731in}{-0.014731in}}%
\pgfpathcurveto{\pgfqpoint{-0.010825in}{-0.018638in}}{\pgfqpoint{-0.005525in}{-0.020833in}}{\pgfqpoint{0.000000in}{-0.020833in}}%
\pgfpathclose%
\pgfusepath{stroke,fill}%
}%
\begin{pgfscope}%
\pgfsys@transformshift{1.332950in}{0.716169in}%
\pgfsys@useobject{currentmarker}{}%
\end{pgfscope}%
\end{pgfscope}%
\begin{pgfscope}%
\pgfpathrectangle{\pgfqpoint{0.626150in}{0.575369in}}{\pgfqpoint{4.650000in}{1.540000in}}%
\pgfusepath{clip}%
\pgfsetbuttcap%
\pgfsetroundjoin%
\pgfsetlinewidth{1.505625pt}%
\definecolor{currentstroke}{rgb}{0.000000,0.000000,0.000000}%
\pgfsetstrokecolor{currentstroke}%
\pgfsetdash{{5.550000pt}{2.400000pt}}{0.000000pt}%
\pgfpathmoveto{\pgfqpoint{1.667750in}{1.811769in}}%
\pgfpathlineto{\pgfqpoint{1.788650in}{1.846969in}}%
\pgfusepath{stroke}%
\end{pgfscope}%
\begin{pgfscope}%
\pgfpathrectangle{\pgfqpoint{0.626150in}{0.575369in}}{\pgfqpoint{4.650000in}{1.540000in}}%
\pgfusepath{clip}%
\pgfsetbuttcap%
\pgfsetroundjoin%
\definecolor{currentfill}{rgb}{0.000000,0.000000,0.000000}%
\pgfsetfillcolor{currentfill}%
\pgfsetlinewidth{1.003750pt}%
\definecolor{currentstroke}{rgb}{0.000000,0.000000,0.000000}%
\pgfsetstrokecolor{currentstroke}%
\pgfsetdash{}{0pt}%
\pgfsys@defobject{currentmarker}{\pgfqpoint{-0.020833in}{-0.020833in}}{\pgfqpoint{0.020833in}{0.020833in}}{%
\pgfpathmoveto{\pgfqpoint{0.000000in}{-0.020833in}}%
\pgfpathcurveto{\pgfqpoint{0.005525in}{-0.020833in}}{\pgfqpoint{0.010825in}{-0.018638in}}{\pgfqpoint{0.014731in}{-0.014731in}}%
\pgfpathcurveto{\pgfqpoint{0.018638in}{-0.010825in}}{\pgfqpoint{0.020833in}{-0.005525in}}{\pgfqpoint{0.020833in}{0.000000in}}%
\pgfpathcurveto{\pgfqpoint{0.020833in}{0.005525in}}{\pgfqpoint{0.018638in}{0.010825in}}{\pgfqpoint{0.014731in}{0.014731in}}%
\pgfpathcurveto{\pgfqpoint{0.010825in}{0.018638in}}{\pgfqpoint{0.005525in}{0.020833in}}{\pgfqpoint{0.000000in}{0.020833in}}%
\pgfpathcurveto{\pgfqpoint{-0.005525in}{0.020833in}}{\pgfqpoint{-0.010825in}{0.018638in}}{\pgfqpoint{-0.014731in}{0.014731in}}%
\pgfpathcurveto{\pgfqpoint{-0.018638in}{0.010825in}}{\pgfqpoint{-0.020833in}{0.005525in}}{\pgfqpoint{-0.020833in}{0.000000in}}%
\pgfpathcurveto{\pgfqpoint{-0.020833in}{-0.005525in}}{\pgfqpoint{-0.018638in}{-0.010825in}}{\pgfqpoint{-0.014731in}{-0.014731in}}%
\pgfpathcurveto{\pgfqpoint{-0.010825in}{-0.018638in}}{\pgfqpoint{-0.005525in}{-0.020833in}}{\pgfqpoint{0.000000in}{-0.020833in}}%
\pgfpathclose%
\pgfusepath{stroke,fill}%
}%
\begin{pgfscope}%
\pgfsys@transformshift{1.667750in}{1.811769in}%
\pgfsys@useobject{currentmarker}{}%
\end{pgfscope}%
\begin{pgfscope}%
\pgfsys@transformshift{1.788650in}{1.846969in}%
\pgfsys@useobject{currentmarker}{}%
\end{pgfscope}%
\end{pgfscope}%
\begin{pgfscope}%
\pgfpathrectangle{\pgfqpoint{0.626150in}{0.575369in}}{\pgfqpoint{4.650000in}{1.540000in}}%
\pgfusepath{clip}%
\pgfsetrectcap%
\pgfsetroundjoin%
\pgfsetlinewidth{1.505625pt}%
\definecolor{currentstroke}{rgb}{0.000000,0.000000,0.000000}%
\pgfsetstrokecolor{currentstroke}%
\pgfsetdash{}{0pt}%
\pgfpathmoveto{\pgfqpoint{5.136650in}{1.987769in}}%
\pgfusepath{stroke}%
\end{pgfscope}%
\begin{pgfscope}%
\pgfpathrectangle{\pgfqpoint{0.626150in}{0.575369in}}{\pgfqpoint{4.650000in}{1.540000in}}%
\pgfusepath{clip}%
\pgfsetbuttcap%
\pgfsetroundjoin%
\definecolor{currentfill}{rgb}{0.000000,0.000000,0.000000}%
\pgfsetfillcolor{currentfill}%
\pgfsetlinewidth{1.003750pt}%
\definecolor{currentstroke}{rgb}{0.000000,0.000000,0.000000}%
\pgfsetstrokecolor{currentstroke}%
\pgfsetdash{}{0pt}%
\pgfsys@defobject{currentmarker}{\pgfqpoint{-0.020833in}{-0.020833in}}{\pgfqpoint{0.020833in}{0.020833in}}{%
\pgfpathmoveto{\pgfqpoint{0.000000in}{-0.020833in}}%
\pgfpathcurveto{\pgfqpoint{0.005525in}{-0.020833in}}{\pgfqpoint{0.010825in}{-0.018638in}}{\pgfqpoint{0.014731in}{-0.014731in}}%
\pgfpathcurveto{\pgfqpoint{0.018638in}{-0.010825in}}{\pgfqpoint{0.020833in}{-0.005525in}}{\pgfqpoint{0.020833in}{0.000000in}}%
\pgfpathcurveto{\pgfqpoint{0.020833in}{0.005525in}}{\pgfqpoint{0.018638in}{0.010825in}}{\pgfqpoint{0.014731in}{0.014731in}}%
\pgfpathcurveto{\pgfqpoint{0.010825in}{0.018638in}}{\pgfqpoint{0.005525in}{0.020833in}}{\pgfqpoint{0.000000in}{0.020833in}}%
\pgfpathcurveto{\pgfqpoint{-0.005525in}{0.020833in}}{\pgfqpoint{-0.010825in}{0.018638in}}{\pgfqpoint{-0.014731in}{0.014731in}}%
\pgfpathcurveto{\pgfqpoint{-0.018638in}{0.010825in}}{\pgfqpoint{-0.020833in}{0.005525in}}{\pgfqpoint{-0.020833in}{0.000000in}}%
\pgfpathcurveto{\pgfqpoint{-0.020833in}{-0.005525in}}{\pgfqpoint{-0.018638in}{-0.010825in}}{\pgfqpoint{-0.014731in}{-0.014731in}}%
\pgfpathcurveto{\pgfqpoint{-0.010825in}{-0.018638in}}{\pgfqpoint{-0.005525in}{-0.020833in}}{\pgfqpoint{0.000000in}{-0.020833in}}%
\pgfpathclose%
\pgfusepath{stroke,fill}%
}%
\begin{pgfscope}%
\pgfsys@transformshift{5.136650in}{1.987769in}%
\pgfsys@useobject{currentmarker}{}%
\end{pgfscope}%
\end{pgfscope}%
\begin{pgfscope}%
\pgfsetrectcap%
\pgfsetmiterjoin%
\pgfsetlinewidth{0.803000pt}%
\definecolor{currentstroke}{rgb}{0.000000,0.000000,0.000000}%
\pgfsetstrokecolor{currentstroke}%
\pgfsetdash{}{0pt}%
\pgfpathmoveto{\pgfqpoint{0.626150in}{0.575369in}}%
\pgfpathlineto{\pgfqpoint{0.626150in}{2.115369in}}%
\pgfusepath{stroke}%
\end{pgfscope}%
\begin{pgfscope}%
\pgfsetrectcap%
\pgfsetmiterjoin%
\pgfsetlinewidth{0.803000pt}%
\definecolor{currentstroke}{rgb}{0.000000,0.000000,0.000000}%
\pgfsetstrokecolor{currentstroke}%
\pgfsetdash{}{0pt}%
\pgfpathmoveto{\pgfqpoint{5.276150in}{0.575369in}}%
\pgfpathlineto{\pgfqpoint{5.276150in}{2.115369in}}%
\pgfusepath{stroke}%
\end{pgfscope}%
\begin{pgfscope}%
\pgfsetrectcap%
\pgfsetmiterjoin%
\pgfsetlinewidth{0.803000pt}%
\definecolor{currentstroke}{rgb}{0.000000,0.000000,0.000000}%
\pgfsetstrokecolor{currentstroke}%
\pgfsetdash{}{0pt}%
\pgfpathmoveto{\pgfqpoint{0.626150in}{0.575369in}}%
\pgfpathlineto{\pgfqpoint{5.276150in}{0.575369in}}%
\pgfusepath{stroke}%
\end{pgfscope}%
\begin{pgfscope}%
\pgfsetrectcap%
\pgfsetmiterjoin%
\pgfsetlinewidth{0.803000pt}%
\definecolor{currentstroke}{rgb}{0.000000,0.000000,0.000000}%
\pgfsetstrokecolor{currentstroke}%
\pgfsetdash{}{0pt}%
\pgfpathmoveto{\pgfqpoint{0.626150in}{2.115369in}}%
\pgfpathlineto{\pgfqpoint{5.276150in}{2.115369in}}%
\pgfusepath{stroke}%
\end{pgfscope}%
\begin{pgfscope}%
\pgftext[x=0.886550in,y=1.807369in,left,base]{\sffamily\fontsize{12.000000}{14.400000}\selectfont \textbf{CRT-6D}}%
\end{pgfscope}%
\begin{pgfscope}%
\pgftext[x=1.277150in,y=1.428969in,left,base]{\sffamily\fontsize{12.000000}{14.400000}\selectfont SOPose \cite{sopose}}%
\end{pgfscope}%
\begin{pgfscope}%
\pgftext[x=1.109750in,y=1.195769in,left,base]{\sffamily\fontsize{12.000000}{14.400000}\selectfont GDRNet \cite{gdr}}%
\end{pgfscope}%
\begin{pgfscope}%
\pgftext[x=1.360850in,y=0.628169in,left,base]{\sffamily\fontsize{12.000000}{14.400000}\selectfont CDPN \cite{cdpn}}%
\end{pgfscope}%
\begin{pgfscope}%
\pgftext[x=1.816550in,y=1.851369in,left,base]{\sffamily\fontsize{12.000000}{14.400000}\selectfont SC6D \cite{SC6D}}%
\end{pgfscope}%
\begin{pgfscope}%
\pgftext[x=4.206650in,y=1.811769in,left,base]{\sffamily\fontsize{12.000000}{14.400000}\selectfont CosyPose \cite{cosypose}}%
\end{pgfscope}%
\end{pgfpicture}%
\makeatother%
\endgroup%

%% file: resources/keypoints.pgf
\begingroup%
\makeatletter%
\begin{pgfpicture}%
\pgfpathrectangle{\pgfpointorigin}{\pgfqpoint{5.411150in}{2.278682in}}%
\pgfusepath{use as bounding box, clip}%
\begin{pgfscope}%
\pgfsetbuttcap%
\pgfsetmiterjoin%
\definecolor{currentfill}{rgb}{1.000000,1.000000,1.000000}%
\pgfsetfillcolor{currentfill}%
\pgfsetlinewidth{0.000000pt}%
\definecolor{currentstroke}{rgb}{1.000000,1.000000,1.000000}%
\pgfsetstrokecolor{currentstroke}%
\pgfsetdash{}{0pt}%
\pgfpathmoveto{\pgfqpoint{0.000000in}{0.000000in}}%
\pgfpathlineto{\pgfqpoint{5.411150in}{0.000000in}}%
\pgfpathlineto{\pgfqpoint{5.411150in}{2.278682in}}%
\pgfpathlineto{\pgfqpoint{0.000000in}{2.278682in}}%
\pgfpathclose%
\pgfusepath{fill}%
\end{pgfscope}%
\begin{pgfscope}%
\pgfsetbuttcap%
\pgfsetmiterjoin%
\definecolor{currentfill}{rgb}{1.000000,1.000000,1.000000}%
\pgfsetfillcolor{currentfill}%
\pgfsetlinewidth{0.000000pt}%
\definecolor{currentstroke}{rgb}{0.000000,0.000000,0.000000}%
\pgfsetstrokecolor{currentstroke}%
\pgfsetstrokeopacity{0.000000}%
\pgfsetdash{}{0pt}%
\pgfpathmoveto{\pgfqpoint{0.626150in}{0.575369in}}%
\pgfpathlineto{\pgfqpoint{5.276150in}{0.575369in}}%
\pgfpathlineto{\pgfqpoint{5.276150in}{2.115369in}}%
\pgfpathlineto{\pgfqpoint{0.626150in}{2.115369in}}%
\pgfpathclose%
\pgfusepath{fill}%
\end{pgfscope}%
\begin{pgfscope}%
\pgfsetbuttcap%
\pgfsetroundjoin%
\definecolor{currentfill}{rgb}{0.000000,0.000000,0.000000}%
\pgfsetfillcolor{currentfill}%
\pgfsetlinewidth{0.803000pt}%
\definecolor{currentstroke}{rgb}{0.000000,0.000000,0.000000}%
\pgfsetstrokecolor{currentstroke}%
\pgfsetdash{}{0pt}%
\pgfsys@defobject{currentmarker}{\pgfqpoint{0.000000in}{-0.048611in}}{\pgfqpoint{0.000000in}{0.000000in}}{%
\pgfpathmoveto{\pgfqpoint{0.000000in}{0.000000in}}%
\pgfpathlineto{\pgfqpoint{0.000000in}{-0.048611in}}%
\pgfusepath{stroke,fill}%
}%
\begin{pgfscope}%
\pgfsys@transformshift{0.626150in}{0.575369in}%
\pgfsys@useobject{currentmarker}{}%
\end{pgfscope}%
\end{pgfscope}%
\begin{pgfscope}%
\pgftext[x=0.626150in,y=0.478146in,,top]{\sffamily\fontsize{12.000000}{14.400000}\selectfont 0}%
\end{pgfscope}%
\begin{pgfscope}%
\pgfsetbuttcap%
\pgfsetroundjoin%
\definecolor{currentfill}{rgb}{0.000000,0.000000,0.000000}%
\pgfsetfillcolor{currentfill}%
\pgfsetlinewidth{0.803000pt}%
\definecolor{currentstroke}{rgb}{0.000000,0.000000,0.000000}%
\pgfsetstrokecolor{currentstroke}%
\pgfsetdash{}{0pt}%
\pgfsys@defobject{currentmarker}{\pgfqpoint{0.000000in}{-0.048611in}}{\pgfqpoint{0.000000in}{0.000000in}}{%
\pgfpathmoveto{\pgfqpoint{0.000000in}{0.000000in}}%
\pgfpathlineto{\pgfqpoint{0.000000in}{-0.048611in}}%
\pgfusepath{stroke,fill}%
}%
\begin{pgfscope}%
\pgfsys@transformshift{1.315039in}{0.575369in}%
\pgfsys@useobject{currentmarker}{}%
\end{pgfscope}%
\end{pgfscope}%
\begin{pgfscope}%
\pgftext[x=1.315039in,y=0.478146in,,top]{\sffamily\fontsize{12.000000}{14.400000}\selectfont 20}%
\end{pgfscope}%
\begin{pgfscope}%
\pgfsetbuttcap%
\pgfsetroundjoin%
\definecolor{currentfill}{rgb}{0.000000,0.000000,0.000000}%
\pgfsetfillcolor{currentfill}%
\pgfsetlinewidth{0.803000pt}%
\definecolor{currentstroke}{rgb}{0.000000,0.000000,0.000000}%
\pgfsetstrokecolor{currentstroke}%
\pgfsetdash{}{0pt}%
\pgfsys@defobject{currentmarker}{\pgfqpoint{0.000000in}{-0.048611in}}{\pgfqpoint{0.000000in}{0.000000in}}{%
\pgfpathmoveto{\pgfqpoint{0.000000in}{0.000000in}}%
\pgfpathlineto{\pgfqpoint{0.000000in}{-0.048611in}}%
\pgfusepath{stroke,fill}%
}%
\begin{pgfscope}%
\pgfsys@transformshift{2.003928in}{0.575369in}%
\pgfsys@useobject{currentmarker}{}%
\end{pgfscope}%
\end{pgfscope}%
\begin{pgfscope}%
\pgftext[x=2.003928in,y=0.478146in,,top]{\sffamily\fontsize{12.000000}{14.400000}\selectfont 40}%
\end{pgfscope}%
\begin{pgfscope}%
\pgfsetbuttcap%
\pgfsetroundjoin%
\definecolor{currentfill}{rgb}{0.000000,0.000000,0.000000}%
\pgfsetfillcolor{currentfill}%
\pgfsetlinewidth{0.803000pt}%
\definecolor{currentstroke}{rgb}{0.000000,0.000000,0.000000}%
\pgfsetstrokecolor{currentstroke}%
\pgfsetdash{}{0pt}%
\pgfsys@defobject{currentmarker}{\pgfqpoint{0.000000in}{-0.048611in}}{\pgfqpoint{0.000000in}{0.000000in}}{%
\pgfpathmoveto{\pgfqpoint{0.000000in}{0.000000in}}%
\pgfpathlineto{\pgfqpoint{0.000000in}{-0.048611in}}%
\pgfusepath{stroke,fill}%
}%
\begin{pgfscope}%
\pgfsys@transformshift{2.692817in}{0.575369in}%
\pgfsys@useobject{currentmarker}{}%
\end{pgfscope}%
\end{pgfscope}%
\begin{pgfscope}%
\pgftext[x=2.692817in,y=0.478146in,,top]{\sffamily\fontsize{12.000000}{14.400000}\selectfont 60}%
\end{pgfscope}%
\begin{pgfscope}%
\pgfsetbuttcap%
\pgfsetroundjoin%
\definecolor{currentfill}{rgb}{0.000000,0.000000,0.000000}%
\pgfsetfillcolor{currentfill}%
\pgfsetlinewidth{0.803000pt}%
\definecolor{currentstroke}{rgb}{0.000000,0.000000,0.000000}%
\pgfsetstrokecolor{currentstroke}%
\pgfsetdash{}{0pt}%
\pgfsys@defobject{currentmarker}{\pgfqpoint{0.000000in}{-0.048611in}}{\pgfqpoint{0.000000in}{0.000000in}}{%
\pgfpathmoveto{\pgfqpoint{0.000000in}{0.000000in}}%
\pgfpathlineto{\pgfqpoint{0.000000in}{-0.048611in}}%
\pgfusepath{stroke,fill}%
}%
\begin{pgfscope}%
\pgfsys@transformshift{3.381706in}{0.575369in}%
\pgfsys@useobject{currentmarker}{}%
\end{pgfscope}%
\end{pgfscope}%
\begin{pgfscope}%
\pgftext[x=3.381706in,y=0.478146in,,top]{\sffamily\fontsize{12.000000}{14.400000}\selectfont 80}%
\end{pgfscope}%
\begin{pgfscope}%
\pgfsetbuttcap%
\pgfsetroundjoin%
\definecolor{currentfill}{rgb}{0.000000,0.000000,0.000000}%
\pgfsetfillcolor{currentfill}%
\pgfsetlinewidth{0.803000pt}%
\definecolor{currentstroke}{rgb}{0.000000,0.000000,0.000000}%
\pgfsetstrokecolor{currentstroke}%
\pgfsetdash{}{0pt}%
\pgfsys@defobject{currentmarker}{\pgfqpoint{0.000000in}{-0.048611in}}{\pgfqpoint{0.000000in}{0.000000in}}{%
\pgfpathmoveto{\pgfqpoint{0.000000in}{0.000000in}}%
\pgfpathlineto{\pgfqpoint{0.000000in}{-0.048611in}}%
\pgfusepath{stroke,fill}%
}%
\begin{pgfscope}%
\pgfsys@transformshift{4.070594in}{0.575369in}%
\pgfsys@useobject{currentmarker}{}%
\end{pgfscope}%
\end{pgfscope}%
\begin{pgfscope}%
\pgftext[x=4.070594in,y=0.478146in,,top]{\sffamily\fontsize{12.000000}{14.400000}\selectfont 100}%
\end{pgfscope}%
\begin{pgfscope}%
\pgfsetbuttcap%
\pgfsetroundjoin%
\definecolor{currentfill}{rgb}{0.000000,0.000000,0.000000}%
\pgfsetfillcolor{currentfill}%
\pgfsetlinewidth{0.803000pt}%
\definecolor{currentstroke}{rgb}{0.000000,0.000000,0.000000}%
\pgfsetstrokecolor{currentstroke}%
\pgfsetdash{}{0pt}%
\pgfsys@defobject{currentmarker}{\pgfqpoint{0.000000in}{-0.048611in}}{\pgfqpoint{0.000000in}{0.000000in}}{%
\pgfpathmoveto{\pgfqpoint{0.000000in}{0.000000in}}%
\pgfpathlineto{\pgfqpoint{0.000000in}{-0.048611in}}%
\pgfusepath{stroke,fill}%
}%
\begin{pgfscope}%
\pgfsys@transformshift{4.759483in}{0.575369in}%
\pgfsys@useobject{currentmarker}{}%
\end{pgfscope}%
\end{pgfscope}%
\begin{pgfscope}%
\pgftext[x=4.759483in,y=0.478146in,,top]{\sffamily\fontsize{12.000000}{14.400000}\selectfont 120}%
\end{pgfscope}%
\begin{pgfscope}%
\pgftext[x=2.951150in,y=0.261295in,,top]{\sffamily\fontsize{12.000000}{14.400000}\selectfont Number of Keypoints}%
\end{pgfscope}%
\begin{pgfscope}%
\pgfsetbuttcap%
\pgfsetroundjoin%
\definecolor{currentfill}{rgb}{0.000000,0.000000,0.000000}%
\pgfsetfillcolor{currentfill}%
\pgfsetlinewidth{0.803000pt}%
\definecolor{currentstroke}{rgb}{0.000000,0.000000,0.000000}%
\pgfsetstrokecolor{currentstroke}%
\pgfsetdash{}{0pt}%
\pgfsys@defobject{currentmarker}{\pgfqpoint{-0.048611in}{0.000000in}}{\pgfqpoint{0.000000in}{0.000000in}}{%
\pgfpathmoveto{\pgfqpoint{0.000000in}{0.000000in}}%
\pgfpathlineto{\pgfqpoint{-0.048611in}{0.000000in}}%
\pgfusepath{stroke,fill}%
}%
\begin{pgfscope}%
\pgfsys@transformshift{0.626150in}{0.575369in}%
\pgfsys@useobject{currentmarker}{}%
\end{pgfscope}%
\end{pgfscope}%
\begin{pgfscope}%
\pgftext[x=0.316851in,y=0.512055in,left,base]{\sffamily\fontsize{12.000000}{14.400000}\selectfont 50}%
\end{pgfscope}%
\begin{pgfscope}%
\pgfsetbuttcap%
\pgfsetroundjoin%
\definecolor{currentfill}{rgb}{0.000000,0.000000,0.000000}%
\pgfsetfillcolor{currentfill}%
\pgfsetlinewidth{0.803000pt}%
\definecolor{currentstroke}{rgb}{0.000000,0.000000,0.000000}%
\pgfsetstrokecolor{currentstroke}%
\pgfsetdash{}{0pt}%
\pgfsys@defobject{currentmarker}{\pgfqpoint{-0.048611in}{0.000000in}}{\pgfqpoint{0.000000in}{0.000000in}}{%
\pgfpathmoveto{\pgfqpoint{0.000000in}{0.000000in}}%
\pgfpathlineto{\pgfqpoint{-0.048611in}{0.000000in}}%
\pgfusepath{stroke,fill}%
}%
\begin{pgfscope}%
\pgfsys@transformshift{0.626150in}{1.088702in}%
\pgfsys@useobject{currentmarker}{}%
\end{pgfscope}%
\end{pgfscope}%
\begin{pgfscope}%
\pgftext[x=0.316851in,y=1.025388in,left,base]{\sffamily\fontsize{12.000000}{14.400000}\selectfont 60}%
\end{pgfscope}%
\begin{pgfscope}%
\pgfsetbuttcap%
\pgfsetroundjoin%
\definecolor{currentfill}{rgb}{0.000000,0.000000,0.000000}%
\pgfsetfillcolor{currentfill}%
\pgfsetlinewidth{0.803000pt}%
\definecolor{currentstroke}{rgb}{0.000000,0.000000,0.000000}%
\pgfsetstrokecolor{currentstroke}%
\pgfsetdash{}{0pt}%
\pgfsys@defobject{currentmarker}{\pgfqpoint{-0.048611in}{0.000000in}}{\pgfqpoint{0.000000in}{0.000000in}}{%
\pgfpathmoveto{\pgfqpoint{0.000000in}{0.000000in}}%
\pgfpathlineto{\pgfqpoint{-0.048611in}{0.000000in}}%
\pgfusepath{stroke,fill}%
}%
\begin{pgfscope}%
\pgfsys@transformshift{0.626150in}{1.602035in}%
\pgfsys@useobject{currentmarker}{}%
\end{pgfscope}%
\end{pgfscope}%
\begin{pgfscope}%
\pgftext[x=0.316851in,y=1.538721in,left,base]{\sffamily\fontsize{12.000000}{14.400000}\selectfont 70}%
\end{pgfscope}%
\begin{pgfscope}%
\pgfsetbuttcap%
\pgfsetroundjoin%
\definecolor{currentfill}{rgb}{0.000000,0.000000,0.000000}%
\pgfsetfillcolor{currentfill}%
\pgfsetlinewidth{0.803000pt}%
\definecolor{currentstroke}{rgb}{0.000000,0.000000,0.000000}%
\pgfsetstrokecolor{currentstroke}%
\pgfsetdash{}{0pt}%
\pgfsys@defobject{currentmarker}{\pgfqpoint{-0.048611in}{0.000000in}}{\pgfqpoint{0.000000in}{0.000000in}}{%
\pgfpathmoveto{\pgfqpoint{0.000000in}{0.000000in}}%
\pgfpathlineto{\pgfqpoint{-0.048611in}{0.000000in}}%
\pgfusepath{stroke,fill}%
}%
\begin{pgfscope}%
\pgfsys@transformshift{0.626150in}{2.115369in}%
\pgfsys@useobject{currentmarker}{}%
\end{pgfscope}%
\end{pgfscope}%
\begin{pgfscope}%
\pgftext[x=0.316851in,y=2.052055in,left,base]{\sffamily\fontsize{12.000000}{14.400000}\selectfont 80}%
\end{pgfscope}%
\begin{pgfscope}%
\pgftext[x=0.261295in,y=1.345369in,,bottom,rotate=90.000000]{\sffamily\fontsize{12.000000}{14.400000}\selectfont Average Recall}%
\end{pgfscope}%
\begin{pgfscope}%
\pgfpathrectangle{\pgfqpoint{0.626150in}{0.575369in}}{\pgfqpoint{4.650000in}{1.540000in}}%
\pgfusepath{clip}%
\pgfsetbuttcap%
\pgfsetroundjoin%
\pgfsetlinewidth{1.505625pt}%
\definecolor{currentstroke}{rgb}{1.000000,0.000000,0.000000}%
\pgfsetstrokecolor{currentstroke}%
\pgfsetdash{{5.550000pt}{2.400000pt}}{0.000000pt}%
\pgfpathmoveto{\pgfqpoint{0.901706in}{0.790969in}}%
\pgfpathlineto{\pgfqpoint{1.177261in}{1.288902in}}%
\pgfpathlineto{\pgfqpoint{1.728372in}{1.365902in}}%
\pgfpathlineto{\pgfqpoint{2.830594in}{1.437769in}}%
\pgfpathlineto{\pgfqpoint{5.035039in}{1.401835in}}%
\pgfusepath{stroke}%
\end{pgfscope}%
\begin{pgfscope}%
\pgfpathrectangle{\pgfqpoint{0.626150in}{0.575369in}}{\pgfqpoint{4.650000in}{1.540000in}}%
\pgfusepath{clip}%
\pgfsetbuttcap%
\pgfsetmiterjoin%
\definecolor{currentfill}{rgb}{1.000000,0.000000,0.000000}%
\pgfsetfillcolor{currentfill}%
\pgfsetlinewidth{1.003750pt}%
\definecolor{currentstroke}{rgb}{1.000000,0.000000,0.000000}%
\pgfsetstrokecolor{currentstroke}%
\pgfsetdash{}{0pt}%
\pgfsys@defobject{currentmarker}{\pgfqpoint{-0.041667in}{-0.041667in}}{\pgfqpoint{0.041667in}{0.041667in}}{%
\pgfpathmoveto{\pgfqpoint{0.000000in}{0.041667in}}%
\pgfpathlineto{\pgfqpoint{-0.041667in}{-0.041667in}}%
\pgfpathlineto{\pgfqpoint{0.041667in}{-0.041667in}}%
\pgfpathclose%
\pgfusepath{stroke,fill}%
}%
\begin{pgfscope}%
\pgfsys@transformshift{0.901706in}{0.790969in}%
\pgfsys@useobject{currentmarker}{}%
\end{pgfscope}%
\begin{pgfscope}%
\pgfsys@transformshift{1.177261in}{1.288902in}%
\pgfsys@useobject{currentmarker}{}%
\end{pgfscope}%
\begin{pgfscope}%
\pgfsys@transformshift{1.728372in}{1.365902in}%
\pgfsys@useobject{currentmarker}{}%
\end{pgfscope}%
\begin{pgfscope}%
\pgfsys@transformshift{2.830594in}{1.437769in}%
\pgfsys@useobject{currentmarker}{}%
\end{pgfscope}%
\begin{pgfscope}%
\pgfsys@transformshift{5.035039in}{1.401835in}%
\pgfsys@useobject{currentmarker}{}%
\end{pgfscope}%
\end{pgfscope}%
\begin{pgfscope}%
\pgfpathrectangle{\pgfqpoint{0.626150in}{0.575369in}}{\pgfqpoint{4.650000in}{1.540000in}}%
\pgfusepath{clip}%
\pgfsetbuttcap%
\pgfsetroundjoin%
\pgfsetlinewidth{1.505625pt}%
\definecolor{currentstroke}{rgb}{0.000000,0.000000,1.000000}%
\pgfsetstrokecolor{currentstroke}%
\pgfsetdash{{5.550000pt}{2.400000pt}}{0.000000pt}%
\pgfpathmoveto{\pgfqpoint{0.901706in}{1.140035in}}%
\pgfpathlineto{\pgfqpoint{1.177261in}{1.720102in}}%
\pgfpathlineto{\pgfqpoint{1.728372in}{1.874102in}}%
\pgfpathlineto{\pgfqpoint{2.830594in}{1.925435in}}%
\pgfpathlineto{\pgfqpoint{5.035039in}{1.822769in}}%
\pgfusepath{stroke}%
\end{pgfscope}%
\begin{pgfscope}%
\pgfpathrectangle{\pgfqpoint{0.626150in}{0.575369in}}{\pgfqpoint{4.650000in}{1.540000in}}%
\pgfusepath{clip}%
\pgfsetbuttcap%
\pgfsetmiterjoin%
\definecolor{currentfill}{rgb}{0.000000,0.000000,1.000000}%
\pgfsetfillcolor{currentfill}%
\pgfsetlinewidth{1.003750pt}%
\definecolor{currentstroke}{rgb}{0.000000,0.000000,1.000000}%
\pgfsetstrokecolor{currentstroke}%
\pgfsetdash{}{0pt}%
\pgfsys@defobject{currentmarker}{\pgfqpoint{-0.041667in}{-0.041667in}}{\pgfqpoint{0.041667in}{0.041667in}}{%
\pgfpathmoveto{\pgfqpoint{0.000000in}{0.041667in}}%
\pgfpathlineto{\pgfqpoint{-0.041667in}{-0.041667in}}%
\pgfpathlineto{\pgfqpoint{0.041667in}{-0.041667in}}%
\pgfpathclose%
\pgfusepath{stroke,fill}%
}%
\begin{pgfscope}%
\pgfsys@transformshift{0.901706in}{1.140035in}%
\pgfsys@useobject{currentmarker}{}%
\end{pgfscope}%
\begin{pgfscope}%
\pgfsys@transformshift{1.177261in}{1.720102in}%
\pgfsys@useobject{currentmarker}{}%
\end{pgfscope}%
\begin{pgfscope}%
\pgfsys@transformshift{1.728372in}{1.874102in}%
\pgfsys@useobject{currentmarker}{}%
\end{pgfscope}%
\begin{pgfscope}%
\pgfsys@transformshift{2.830594in}{1.925435in}%
\pgfsys@useobject{currentmarker}{}%
\end{pgfscope}%
\begin{pgfscope}%
\pgfsys@transformshift{5.035039in}{1.822769in}%
\pgfsys@useobject{currentmarker}{}%
\end{pgfscope}%
\end{pgfscope}%
\begin{pgfscope}%
\pgfsetrectcap%
\pgfsetmiterjoin%
\pgfsetlinewidth{0.803000pt}%
\definecolor{currentstroke}{rgb}{0.000000,0.000000,0.000000}%
\pgfsetstrokecolor{currentstroke}%
\pgfsetdash{}{0pt}%
\pgfpathmoveto{\pgfqpoint{0.626150in}{0.575369in}}%
\pgfpathlineto{\pgfqpoint{0.626150in}{2.115369in}}%
\pgfusepath{stroke}%
\end{pgfscope}%
\begin{pgfscope}%
\pgfsetrectcap%
\pgfsetmiterjoin%
\pgfsetlinewidth{0.803000pt}%
\definecolor{currentstroke}{rgb}{0.000000,0.000000,0.000000}%
\pgfsetstrokecolor{currentstroke}%
\pgfsetdash{}{0pt}%
\pgfpathmoveto{\pgfqpoint{5.276150in}{0.575369in}}%
\pgfpathlineto{\pgfqpoint{5.276150in}{2.115369in}}%
\pgfusepath{stroke}%
\end{pgfscope}%
\begin{pgfscope}%
\pgfsetrectcap%
\pgfsetmiterjoin%
\pgfsetlinewidth{0.803000pt}%
\definecolor{currentstroke}{rgb}{0.000000,0.000000,0.000000}%
\pgfsetstrokecolor{currentstroke}%
\pgfsetdash{}{0pt}%
\pgfpathmoveto{\pgfqpoint{0.626150in}{0.575369in}}%
\pgfpathlineto{\pgfqpoint{5.276150in}{0.575369in}}%
\pgfusepath{stroke}%
\end{pgfscope}%
\begin{pgfscope}%
\pgfsetrectcap%
\pgfsetmiterjoin%
\pgfsetlinewidth{0.803000pt}%
\definecolor{currentstroke}{rgb}{0.000000,0.000000,0.000000}%
\pgfsetstrokecolor{currentstroke}%
\pgfsetdash{}{0pt}%
\pgfpathmoveto{\pgfqpoint{0.626150in}{2.115369in}}%
\pgfpathlineto{\pgfqpoint{5.276150in}{2.115369in}}%
\pgfusepath{stroke}%
\end{pgfscope}%
\end{pgfpicture}%
\makeatother%
\endgroup%

%% file: sections/conclusion.tex
\section{Conclusion}
\label{sec:Conclusion}

We have proposed a novel approach to 6d object pose estimation \Ours~ based on iterative pose refinements. The input to our refinement modules is a set of feature vectors OSKFs, sampled from feature pyramid at the location of known keypoints, 2D projected using a coarse pose. OSKFs are the representation of the pose offset representation and are fed into an OSKF-PoseTransformer  to extract the refined pose. We achieve state of the art on multiple datasets while being at least $2\times$ faster than similar methods.

For future directions, we are looking to extend \Ours~ into category level object pose estimation, where specific keypoints cannot be used.

\textbf{AcknowledgementS.} This work is in part sponsored by KAIA grant (22CTAP-C163793-02, MOLIT), NST grant (CRC 21011, MSIT), KOCCA grant (R2022020028, MCST) and the Samsung Display corporation.